\documentclass[journal, table, xcdraw, twoside]{IEEEtran}

\IEEEoverridecommandlockouts
\usepackage{amsmath,amssymb,amsfonts}
\usepackage{color} 
\usepackage{graphicx} 
\usepackage[utf8]{inputenc}
\usepackage{lipsum}  
\usepackage{multirow} 
\usepackage{units}
\usepackage{textcomp} 
\usepackage{tikz}
\usepackage{pgfplots}
\usepackage{balance}
\usepackage{url}
\usepackage[skip=0pt,font=footnotesize]{caption}
\usepackage[pdftex]{hyperref}
\usepackage{adjustbox}
\usepackage{booktabs}
\usepackage{cite}
\usepackage{float}
\usepackage[normalem]{ulem}
\usepackage{cancel}
\usepackage{tabularx}

\newcolumntype{C}{>{\centering\arraybackslash}X}
\newcolumntype{x}[1]{>{\centering\let\newline\\\arraybackslash\hspace{0pt}}p{#1}}

\DeclareMathOperator{\st}{subject \ to:} 

\DeclareUnicodeCharacter{2212}{−}
\usepgfplotslibrary{groupplots,dateplot}
\usetikzlibrary{patterns,shapes.arrows, matrix, positioning}
\pgfplotsset{compat=1.16}
\pgfdeclarelayer{bg}    %
\pgfsetlayers{bg,main}

\NewDocumentCommand{\evalat}{sO{\big}mm}{%
  \IfBooleanTF{#1}
   {\mleft. #3 \mright|_{#4}}
   {#3#2|_{#4}}%
}

\newcommand{\wfr}[0]{\ensuremath{W}} %
\newcommand{\bfr}[0]{\ensuremath{B}} %

\newcommand{\bm}[1]{\boldsymbol{#1}}
\newcommand{\mat}[1]{\begin{bmatrix}#1\end{bmatrix}}
\newcommand{\rom}[1]{(\expandafter{\romannumeral #1\relax})}

\newcommand{\change}[1]{#1}
\newcommand{\specialchange}[1]{#1}

\newcommand{\comment}[1]{}

\newcommand{\rebuttal}[1]{#1}
\newcommand{\final}[1]{#1}

\definecolor{somegray}{rgb}{0.5, 0.5, 0.5}
\newcommand{\darkgrayed}[1]{\textcolor{somegray}{#1}}
\makeatletter
\newcommand*\titleheader[1]{\gdef\@titleheader{#1}}
\AtBeginDocument{%
  \let\st@red@title\@title
  \def\@title{%
    \vskip-2em
    \bgroup\normalfont\large\centering\@titleheader\par\egroup
    \vskip0.5em\st@red@title}
}
\makeatother

\titleheader{\darkgrayed{This paper has been accepted for publication in the
IEEE Transactions on Robotics (T-RO),\,2024. \textcopyright IEEE}}

\graphicspath{{figures/}}
\title{Autonomous Drone Racing: A Survey}
\author{Drew Hanover$^{1}$, Antonio Loquercio$^{3}$, Leonard Bauersfeld$^{1}$, Angel Romero$^{1}$, Robert Penicka$^{2}$, \\ Yunlong Song$^{1}$, Giovanni Cioffi$^{1}$,  Elia Kaufmann$^{1}$ and Davide Scaramuzza$^{1}$}
\makeatletter
\g@addto@macro\@maketitle{
  \captionsetup{type=figure}\setcounter{figure}{0}
  \vspace*{12pt}
  \includegraphics[width=1\linewidth]{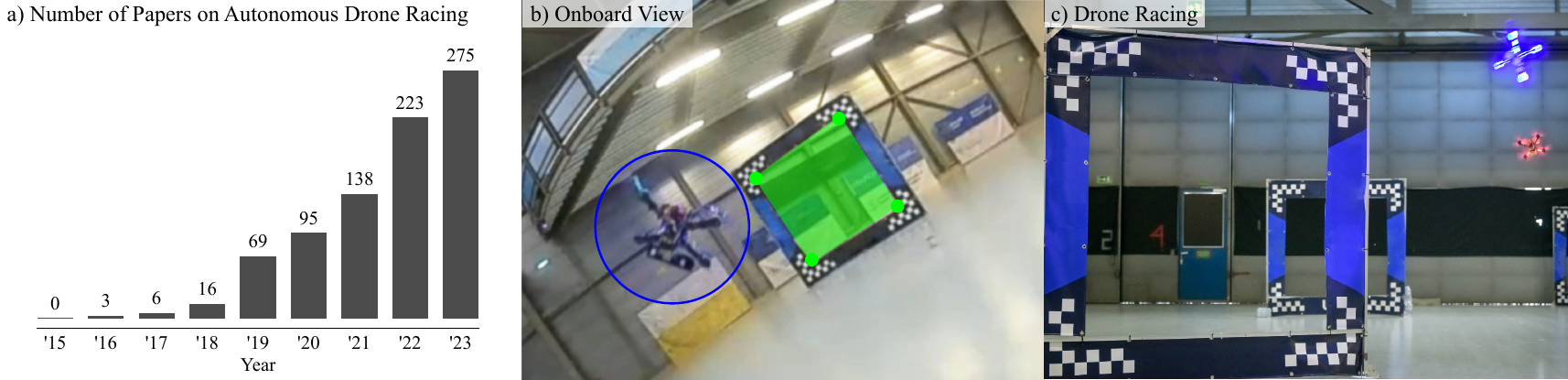}
    \vspace*{-3pt}
	\captionof{figure}{Drone racing is a sport rapidly gaining popularity where opponents compete on a pre-defined race track consisting of a series of gates.
    Autonomous drone racing research aims to build algorithms that can outperform human pilots in such competitions.
    a) The task of autonomous drone racing has gained a substantial amount of interest from the research community in the last few years, as indicated by the increasing number of related publications per year, \change{as evidenced by a google-scholar search for ``autonomous drone racing''.} 
    b) Autonomous drones rely on visual and inertial sensors to estimate their own states, as well as their opponents' states.
    c) Agile maneuvers are required to overtake opponents and win the race.
    }
    \label{fig:sdd_2022}
	\vspace{-9pt}
        \thanks{
        $^{1}$D. Hanover, L. Bauersfeld, A. Romero, Y. Song, G. Cioffi, E. Kaufmann and D. Scaramuzza are with the Robotics and Perception Group, University of Zurich, Switzerland (\protect\url{http://rpg.ifi.uzh.ch}).
        $^{2}$R. Penicka is with the Multi-robot Systems Group, Czech Technical University in Prague, Czech Republic.
        $^{3}$A. Loquercio is with UC Berkeley.
        This work was supported by the Swiss National Science Foundation (SNSF) through the National Centre of Competence in Research (NCCR) Robotics, the Czech Science Foundation (GACR) under research projects No. 23-06162M, the European Union’s Horizon 2020 Research and Innovation Programme under grant agreement No.~871479 (AERIAL-CORE), and the European Research Council (ERC) under grant agreement No.~864042 (AGILEFLIGHT).
        }
	\vspace{-9pt}
}
\makeatother
\date{June 2022}

\begin{document}
\maketitle
\thispagestyle{empty}

\begin{abstract}
Over the last decade, the use of autonomous drone systems for surveying, search and rescue, or last-mile delivery has increased exponentially.
With the rise of these applications comes the need for highly robust, safety-critical algorithms that can operate drones in complex and uncertain environments.  
Additionally, flying fast enables drones to cover more ground, increasing productivity and further strengthening their use case. 
One proxy for developing algorithms used in high-speed navigation is the task of autonomous drone racing, where researchers program drones to fly through a sequence of gates and avoid obstacles as quickly as possible using onboard sensors and limited computational power.
Speeds and accelerations exceed over \unit[80]{kph} and \unit[4]{g}, respectively, raising significant challenges across perception, planning, control, and state estimation.
To achieve maximum performance, systems require real-time algorithms that are robust to motion blur, high dynamic range, model uncertainties, aerodynamic disturbances, and often unpredictable opponents.
This survey covers the progression of autonomous drone racing across model-based and learning-based approaches.
We provide an overview of the field, its evolution over the years, and conclude with the biggest challenges and open questions to be faced in the future. 
\end{abstract}
\vspace*{-6pt}

\section{Introduction}\label{sec:introduction}

Throughout history, humans have been obsessed with racing competitions, where physical and mental fitness are put to the test.
The earliest mention of a formal race dates all the way back to 3000 BC in ancient Egypt, where the Pharaoh was thought to have run a race at the Sed festival to demonstrate his physical fitness, indicating his ability to rule over the kingdom~\cite{wilkinson2002early,arab_2017}.
As time has progressed, humans have moved from racing on-foot to using chariots, cars, planes, and more recently quadcopters\cite{Betz2022_RacingSurvey}. 
Although the vessel frequently changes, one thing that has remained constant since the early days of racing has been the recurring theme of using the task as a catalyst for scientific and engineering development.
Recently, we have seen a push to remove humans from the loop, automating the highly complex task of racing in order to push vehicle performance beyond what humans can achieve~\cite{Song23Reaching, kaufmann23champion}.

\subsection{Why Autonomous Drone Racing?}
Drone racing is a popular sport with high-profile international competitions. 
In a traditional drone race, each vehicle is controlled by a human pilot, who receives a first-person-view~(FPV) live stream from an onboard camera and flies the drone via a radio transmitter.
An onboard image from the drone can be seen in Fig. \ref{fig:sdd_2022}b.
\change{Having access to an FPV feed sets drone racing apart from remote-controlled fixed-wing aircraft racing, where pilots typically control the vehicle in a line-of-sight fashion.}
Human drone pilots need years of training to master the advanced navigation and control skills required to succeed in international competitions.
Such skills would also be valuable to autonomous systems that must fly through complex environments in applications such as disaster response, aerial delivery, and inspection of complex structures.
For example, automating inspection tasks can save lives while being more productive than manual inspection.
According to a recent survey on unmanned aerial vehicle (UAV) use in bridge inspection~\cite{ameli2022impact}, most drones used for inspection tasks rely on GPS navigation with the biggest limiting factor on inspection efficiency being the drones' endurance and mobility.
Additionally, the most popular drones used for surveying by several US Departments of Transportation are not fully autonomous and require expert human pilots~\cite{ameli2022impact}. 
\change{In these applications, an increase in autonomy and operational speed will offer gains in utility as faster flight increases the operating radius achievable with a given battery~\cite{bauersfeld22Range}. Drone racing research has made significant progress in bringing the skills of autonomous drones closer to those of professional human pilots~\cite{kaufmann23champion}. This required advances on all parts of the flight stack, i.e., estimation, planning, control, and hardware~\cite{Foehn22Agi}, which we cover in length in this survey. However, several challenges remain to bridge the gap between drone racing and real-world applications, such as safety~\cite{brunke2022safe} and generalization over tasks and environments.
}

\begin{figure*}[t!]
    \centering
    \includegraphics{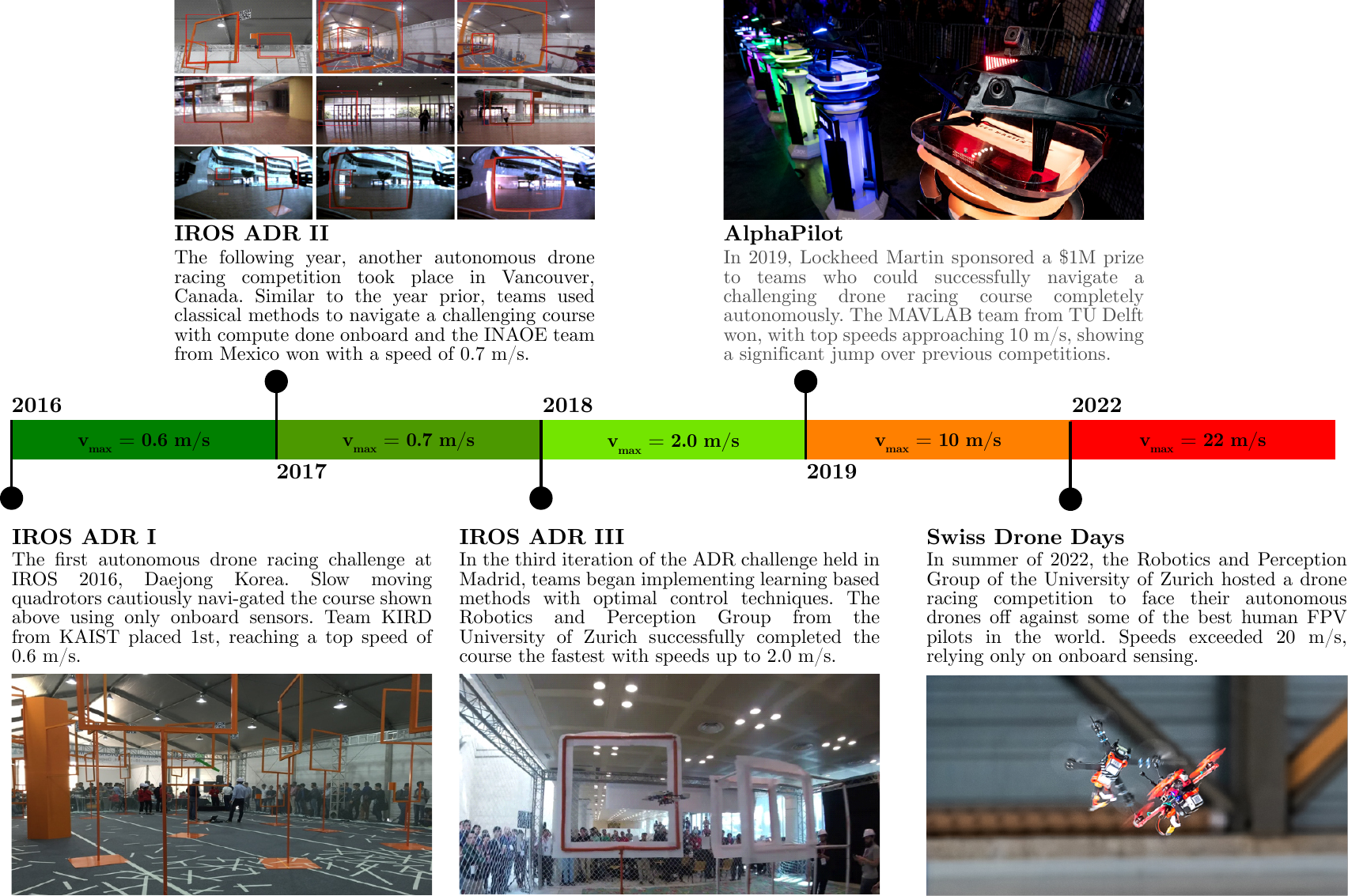}
    \hspace{10mm}
    \caption{History of drone racing competitions that use real drones for navigating the race track, IROS ADR II photo credit~\cite{jung2018direct}. 
    }
    \label{fig:timeline}
\end{figure*}

\change{
Over the last five years, several projects have been launched to encourage rapid progress within the field, such as DARPA's Fast Lightweight Autonomy (FLA) ~\cite{mohta2018fast}, European Research Council's AgileFlight~\cite{ERC2021agileflight}, and the AutoAsses\final{s} project~\cite{AutoAsses}.
With million-dollar funding for each of these projects and significant commercial potential, a large incentive exists for researchers and entrepreneurs alike to achieve autonomous operation in GNSS-denied and confined critical infrastructure.}
Competitions such as the IROS'16-19 Autonomous Drone Racing series \cite{moon2019challenges}, NeurIPS 2019's Game of Drones~\cite{microsoft2019gameofdrones}, and the 2019 AlphaPilot Challenge~\cite{foehn2022alphapilot,de2021artificial} provided further opportunity for researchers to compare their methodologies against one another in a competitive fashion. 
A depiction of the progress made from these competitions can be seen in Fig.~\ref{fig:timeline}. \final{However, we are far from having solved autonomous drone racing\textemdash a notion reflected by the recently announced competition scheduled for 2025 and to be hosted by the Abu Dhabi Autonomous Racing League~\cite{a2rl2024droneracing}.}

Drone racing is a challenging benchmark that can help researchers to gauge progress on complex perception, planning, and control algorithms.
Autonomous drones in a racing setting must be able to perceive, reason, plan, and act on the tens of milliseconds scale, all onboard a computationally limited platform.
Apart from being very challenging to solve, the drone racing task offers a single measure of the progress of the state-of-the-art in autonomous flying robotics: lap time.
Solving this problem requires efficient, lightweight algorithms to provide optimal decision and control behaviors in real-time.
\change{Just a few years back, it was unclear whether such a problem was feasible in the first place, even given perfect information. Drone racing research has advanced much since its early stages~\cite{moon2017iros}. Such advances required radically new algorithmic ideas, e.g., training learning-based sensorimotor controllers only in simulation, together with engineering advances to the platform and the overall system~\cite{Foehn22Agi}.
Now that superhuman performance has been achieved~\cite{kaufmann23champion} (despite being in controlled conditions), we predict that more work will be done on safety and generalization over tasks and environments to bridge the gap between drone racing and real-world applications.
This research effort is already evident today, as shown by the increasing number of papers in the field over the years (Fig.~\ref{fig:sdd_2022}a).}

This is the first survey on the state of the art in autonomous drone racing.
This overview will be useful to researchers who wish to make connections between existing works, learn about the strengths and weaknesses of current and past approaches, and identify directions moving forward which should progress the field in a meaningful way.

\subsection{Task Specification}
The drone racing task is to fly a quadrotor through a sequence of gates in a given order in minimum time while avoiding collisions.
Humans are astonishingly good at this task, flying at speeds well over \unit{100}{kph} with only a first-person view camera as their sensory input.
Beyond this, expert pilots can adapt to new race tracks quickly in a matter of minutes, however, the sensorimotor skills required by professional drone pilots take years of training to acquire.

For an autonomous drone to successfully complete this task, it must be able to detect opponents and waypoints along the track, calculate their location and orientation in 3-dimensional space, and compute an action that enables navigation through the track as quickly as possible while still controlling a highly nonlinear system at the limits.
This is challenging in three different aspects: Perception, Planning, and Control.
Poor design in any of these aspects can make the difference between winning or losing the race, which can be decided by less than a tenth of a second.

The paper is structured as follows:
First, the modeling procedure of the drone, including aerodynamics, batteries, motors, cameras, and the system nonlinearities, are discussed in detail in Sect.~\ref{sec:modeling}.
A classical robotics pipeline is then introduced in Sect.~\ref{ClassicalPipeline}, with a deep dive into literature relevant to agile flight split into Perception, Planning, and Control subsections. \change{All of these components are equally important because, at the edge of a drone's performance envelope, all parts\textemdash perception, state-estimation, planning, and control\textemdash need to meticulously work together.}
Next, we delve into learning-based methods for Perception, Planning, and Control which rely on recent advancements from the machine learning community in Sect.~\ref{Learning-based-Approaches}. 
Then, a discussion of the development of simulation tools that can enable rapid development for agile flight applications in Sect.~\ref{sec:simulators}.
A history of drone racing competitions and the methods used for each are included in Sect.~\ref{sec:competitions}.
Next, a summary of open source code bases, hardware platforms, and datasets for researchers is provided in Sect.~\ref{sec:datasets}.
Finally, a forward-looking discussion on the Opportunities and Challenges for future researchers interested in autonomous drone racing in Sect.~\ref{sec:challenges}.

\section{Drone Modeling}\label{sec:modeling}

To further advance research on fast and agile flight, it is important to have accurate models that capture the complex nonlinear dynamics of multicopter vehicles at the limit of their performance envelope.

This section reviews different dynamics modeling approaches from classic, first-principles modeling to pure data-driven models in the context of drone racing. 
For the vehicle dynamics, the key aspects that need to be modeled are the kinematics, aerodynamics, the electric motors, and the battery. 
In addition to the vehicle dynamics models discussed in this section, many difficulties for autonomous drone racing models are introduced by the onboard sensors, whose characteristics need to be modeled. For example, IMUs are subject to bias and noise, and the intrinsic as well as extrinsic parameters of onboard sensors change over time as hard crashes may lead to miscalibration.

\subsection{Kinematics}
Typically, the vehicle is assumed to be a 6 degree-of-freedom rigid body of mass $m$ with a (diagonal) inertia matrix $\bm{J}=\mathrm{diag}(J_x, J_y, J_z)$.
Given a dynamic state $\bm{x} \in \mathbb{R}^{17}$ the equations describing its evolution in time are commonly written as:
\vspace*{-6pt}
\begin{align}
\small
\label{eq:3d_quad_dynamics}
\dot{\bm{x}} = f(\bm{x}, \bm{u}) =
\begin{bmatrix}
\dot{\bm{p}}_{\wfr\bfr} \\[6pt]  
\dot{\bm{q}}_{\wfr\bfr} \\[6pt]
\dot{\bm{v}}_{\wfr} \\[6pt]
\dot{\boldsymbol\omega}_\bfr \\[6pt]
\dot{\boldsymbol\Omega}
\end{bmatrix} = 
\begin{bmatrix}
\bm{v}_\wfr \\[6pt]
\bm{q}_{\wfr\bfr} \mat{0 \\ \bm{\omega}_\bfr/2} \\[6pt]
\frac{1}{m} \big(\bm{q}_{\wfr\bfr} \odot \bm f\big) +\bm{g}_\wfr  \\[6pt]
\bm{J}^{-1}\big( \boldsymbol \tau - \boldsymbol\omega_\bfr \times \bm{J}\boldsymbol\omega_\bfr \big) \\[6pt]
\frac{1}{\tau_\Omega} (\boldsymbol\Omega_\text{ss} - \boldsymbol{\Omega})
\end{bmatrix} \; ,
\end{align}
where $\bm{p}_{\wfr\bfr} \in \mathbb{R}^{3}$ is the position of the center of mass given in the world frame,  $\bm{q}_{\wfr\bfr} \in SO(3)$ is a quaternion defining the rotation of the body frame relative to the world (vehicle attitude), $\bm{v}_{\wfr} \in \mathbb{R}^{3}$ is the velocity of the vehicle in the world frame, $\boldsymbol{\omega}_{\bfr} \in \mathbb{R}^{3}$ are the bodyrates of the vehicle, $\boldsymbol\Omega \in \mathbb{R}^{4}$ are the motor speeds, $\bm{g}_\wfr= [0, 0, \unit[9.81]{m/s^2}]^\intercal$ denotes earth's gravity, and $\bm{u} \in \mathbb{R}^4$ is the input. Depending on the control modality, the input can be single rotor thrusts or collective thrust and body rates.
In this setting, the task of the model is to calculate the total force $\bm f$ and total torque $\boldsymbol \tau$ that acts on the drone as accurately as possible. 
Note the quaternion-vector product denoted by $\odot$ representing a rotation of the vector by the quaternion as in $\bm{q} \odot \bm{f} = \bm{q} \cdot [0, \bm{f}^\intercal]^\intercal \cdot \bar{\bm{q}}$, where $\bar{\bm{q}}$ is the quaternion's conjugate.
Those forces and torques, collectively referred to as wrench, are determined by the aerodynamics of the platform as well as the vehicles' actuators, e.g. the propellers.

\subsection{Aerodynamics}
This section discusses the different approaches to modeling the aerodynamics of the drone and its propellers. 
The most widely used modeling assumption is that the propeller thrust and drag torque are proportional to the square of the rotational speed~\cite{mahony2012multirotor, shah2018airsim, song2020flightmare, furrer2016rotors, meyer2012comprehensive} and that the body drag is negligible. 
These assumptions quickly break down at the high speeds encountered in drone racing as this model neglects (a) linear rotor drag~\cite{prouty1995helicopter, faessler2017differential}, (b) dynamic lift~\cite{prouty1995helicopter}, (c) rotor-to-rotor~\cite{yoon2018cfd, diaz2018cfd, yoon2017cfd}, (d) rotor-to-body~\cite{yoon2018cfd, diaz2018cfd, yoon2017cfd} interactions and (e) aerodynamic body drag \cite{faessler2017differential, diaz2018cfd}. 
 
The accuracy of the propeller model can be improved by leveraging blade-element-momentum theory, where the propeller is modeled as a rotating wing.
Such first-principle approaches~\cite{gill2017propeller, gill2019computationally, khan2013toward, hoffmann2007quadrotor, bangura2017thrustcontrol} have been shown to provide very accurate models of the wrench generated by a single propeller as they properly capture effects (a) and (b). 
Implemented efficiently, a Blade Element Momentum (BEM) model can be run in  real-time~\cite{bauersfeld2021neurobem} and has been successfully used to test algorithms in simulation~\cite{kaufmann2022benchmark, penicka22RALmintimeplanning}.

Accounting for the remaining open points (c)-(e), the aerodynamics of the drone body as well as any interaction effects need to be calculated, which requires a full Computational Fluid Dynamics (CFD) simulation~\cite{ventura2018high,luo2015novel, diaz2018cfd, yoon2017cfd, yoon2018cfd}. Due to the extreme computational demands, this is impractical in drone racing.
To still get close to the accuracy of CFD methods while retaining the computational simplicity of the previously mentioned methods, data-driven approaches are employed~\cite{bauersfeld2021neurobem, torrente2021datadriven, sun2019quadrotor, bansal2016learning, punjani2015deep, shi2019neurallander}.
In the early works~\cite{bansal2016learning, punjani2015deep}, the whole vehicle dynamics model was learned from data. In a similar fashion~\cite{sun2019quadrotor} uses a combination of polynomials\textemdash identified from wind-tunnel flight data\textemdash to represent the vehicle dynamics.
In~\cite{bauersfeld2021neurobem, torrente2021datadriven}, it has been shown that higher modeling accuracies can be achieved when combining a first-principle model with a data-driven component.
Such a combination of first-principle and data-driven models also leads to improved generalization performance, as shown in~\cite{bauersfeld2021neurobem}, which combines a BEM model with a temporal convolutional network~\cite{oord2016wavenet} to regress the residual wrench. 
\final{Recently, a similar hybrid modeling approach has been applied to moving-horizon estimation~\cite{wang2024neuromhe}.}

\subsection{Motor and Battery Models}
The previous section outlines different approaches to how the aerodynamic wrench can be estimated based on the state of the vehicle. 
However, for all such models, the rotational speed of the propeller is assumed to be known.
On most multicopters, the motors are not equipped with closed-loop motor speed control but are controlled by a \change{`throttle'} command, which controls the duty cycle of a PWM (pulse-width-modulation) signal applied to the motors. 
The actual rotational speed that the motor achieves is a function of the throttle command as well as other parameters such as the battery voltage and the drag torque of the rotor~\cite{bauersfeld22Range}.
Therefore, in order to have a dynamics model for the motors, we need a model of the battery to calculate the voltage applied to the motors.
Most literature on battery modeling relies on so-called Peukert models~\cite{1897:Peukert}, but for lithium-polymer batteries in drone racing, this is hardly applicable because the battery discharge current often exceeds \unit[100]{A} (e.g. \unit[50-100]{C})~\cite{2015:Galushkin, 2020:Galushkin}. 
Graybox battery models for the voltage that are based on a one-time-constant (OTC) equivalent circuit~\cite{2016:Zhang, 2020:Zhang} are much more suitable for drone racing tasks as shown in~\cite{bauersfeld22Range}, because they are applicable to the extremely high loads experienced during a racing scenario.
In combination with either a polynomial or a constant-efficiency motor model, such OTC models can be used to accurately simulate the battery voltage during agile flight~\cite{bauersfeld22Range}. 
Given a simulation of the battery voltage, one can measure the performance characteristics of a given motor-propeller combination to determine the mapping of throttle command and voltage to resulting steady-state propeller speed $\Omega_\text{ss}$.
When the highest model fidelity is desired, a more sophisticated motor simulation~\cite{bicego2020nmpcmotormodel} can further improve the accuracy, which can be desirable if the controller directly outputs single-rotor thrusts instead of the more commonly used collective-thrust and body rates control modality.

\subsection{Camera and IMU Modeling}
Drone racing pushes not only the mechanical and electrical components of drones to their limits, but is also highly demanding in terms of sensor performance. For an in-depth overview of the many different sensor options for drone racing the reader is referred to~\cite{perez2021onboard}. 
The most common sensors aboard autonomous drones are monocular or stereo cameras combined with IMUs (inertial measurement units) thanks to their low cost, low weight, and mechanical robustness. 

\change{For vision-based drone racing, having an accurate simulation of the perception pipeline is critical for validation and controller development. In terms of modeling and simulation of the camera, it is common to use a pinhole model ~\cite{scaramuzza2011visual} and estimate the focal length, image center, and distortion parameters from measurements. Combined with accurate information on how far the camera is displaced from the center of gravity of the vehicle, this allows simulating observations. Either low-level sensory observations (e.g. images) are simulated using a rendering engine~\cite{song2020flightmare, Guerra2019flightgoggles} or more abstract visual features (e.g. landmark positions) are simulated using the projection equations.}

\change{In the context of using a simulation to test approaches before attempting real-world deployment, an accurate model of the IMU characteristics is important, as the bias and noise strongly influence the performance of many methods.} 
The IMU intrinsic calibration estimates the noise characteristic of the sensor.
The camera-IMU extrinsic calibration estimates the relative position and orientation of the two sensors as well as the time offset. Kalibr~\cite{rehder2016extending} is a widespread tool to perform these calibrations.

However, the biggest source of measurement error of the inertial sensors onboard a drone are not the sensors themselves but the strong high-frequency vibrations introduced by the fast-spinning propellers. The vibrations lead to aliasing effects on the IMU measurements and introduce additional motion blur on the camera images.
The structural vibrations and their effect on the measurements are extremely difficult to model and correct for. Therefore, a suitable hardware design is imperative which dampens the mount of the camera and the IMU with respect to the vehicle frame.

\section{Classical Perception, Planning, and Control Pipeline} \label{ClassicalPipeline}
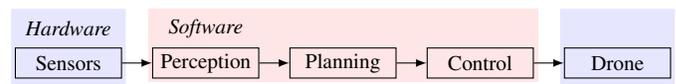
\begin{figure}[ht]
    \centering
    \begin{tikzpicture}[>=latex, font=\footnotesize]
    \tikzstyle{tb}=[minimum width = 1.4cm, text width = 1.2cm, minimum height = 0.4cm, align = center, draw, inner sep = 1pt, inner xsep=3pt];
    \newcommand\dx{0.4cm};
    \newcommand\dy{0.05cm};
    \newcommand\shift{0.05cm};
     \node [tb] (a) {Sensors};
     \node [tb, draw = none, above = \dy of a] (a2) {\emph{Hardware}};
     \node [tb, right = \dx of a] (b) {Perception};
     \node [tb, draw = none, above = \dy of b] (b2) {\emph{Software}};
     \node [tb, right = \dx of b] (c) {Planning};
     \node [tb, right = \dx of c] (d) {Control};
     \node [tb, right = \dx of d] (e) {Drone};
     \node [tb, draw = none, above = \dy of e] (e2) {};
     \draw [->] (a) -- (b);
     \draw [->] (b) -- (c);
     \draw [->] (c) -- (d);
     \draw [->] (d) -- (e);
     \begin{pgfonlayer}{bg}    %
         \draw [draw=none, fill=red!10!white] ([xshift = -\shift, yshift=\shift]b2.north west) rectangle ([xshift=\shift, yshift=-\shift]d.south east);
         \draw [draw = none, fill=blue!10!white] ([xshift = -\shift, yshift=\shift]a2.north west) -| ([xshift=\shift, yshift=-2*\shift]a.south east) -| ([xshift=-\shift, yshift=\shift]e2.north west) -| ([xshift=\shift, yshift=-4*\shift]e.south east) -| ([xshift=\shift, yshift=-4*\shift]a.south east) -| cycle;
    \end{pgfonlayer}
    \end{tikzpicture}
    \vspace*{-8pt}
    \caption{Architecture 1: A classic architecture for an autonomous system programmed using model-based approaches}
    \label{fig:traditionalPipeline}
\end{figure}

Since the inception of mobile robotics, a common architecture has been primarily used to achieve autonomous navigation capabilities across various systems.
In a traditional robotics software stack, the navigation task is broken into three main components: Perception, Planning, and Control.
A diagram of this architecture can be seen in Fig. \ref{fig:traditionalPipeline}.
This section covers recent research in these areas relating specifically to agile flight and autonomous drone racing.
All approaches detailed in this section rely on first principles modeling and optimization techniques.

\subsection{Perception}
\label{sec:classic_perception}

The perception block estimates the vehicle state and perceives the environment using onboard sensors. 
The most common solution for state estimation of flying vehicles is visual-inertial odometry (VIO), thanks to its low cost and low weight requirements.
VIO uses camera and IMU measurements to estimate the state $\hat{\bm{x}}$ (position, orientation, and velocity) of the drone platform.
The inertial measurements are integrated to obtain relative position, orientation, and velocity estimates in a short time, e.g., between two camera images.
However, the integration for a longer time, e.g., a few seconds, accumulates large drift due to scale factor errors, axis misalignment errors, and time-varying biases~\cite{yang2020online} that commonly affect off-the-self IMU measurements.
The camera measurements provide rich information about the environment at a lower rate, usually around \unit[30]{Hz}, than IMU measurements.
Unlike the IMU measurements, the camera measurements are affected by environmental conditions.
The quality of information they provide for state estimation degrades in the case of poor illumination conditions, textureless scenes, and motion blur.
For this reason, the camera and inertial measurements complement each other and are the standard choice for state estimation of flying vehicles \cite{zhang2019encyclopedia}. 
In this section, we first give an overview of VIO with a focus on the methods that can be applied for online state estimation of \rebuttal{a racing drone}.
\rebuttal{Second, we give an overview of possible additional sensor modalities that integrated into the classical VIO pipeline have the potential to improve state estimation at high speed.}
Third, we conclude with a discussion on the application of classical VIO methods to drone racing.

\subsubsection{VIO}

VIO is the most common solution for state estimation of aerial vehicles~\cite{zhang2019encyclopedia} using 
only onboard sensing and computing, thanks to its favorable trade-off between accuracy and computational requirements.
VIO algorithms usually comprise two main blocks: the frontend and the backend.

The frontend uses camera images to estimate the motion of the sensor.
Two main approaches exist in the literature: direct methods~\cite{engel2017direct, bloesch2015robust} and feature-based methods~\cite{mourikis2007multi, leutenegger2015keyframe, qin2018vins}.
Direct methods work directly on the raw pixel intensities.
These methods commonly extract image patches and estimate the camera trajectory by tracking the motion of such patches through consecutive images.
The tracking is achieved by minimizing a photometric error defined on the raw pixel intensities~\cite{engel2017direct}.
\final{This tracking method is particularly interesting for drone racing because of its robustness in featureless scenarios.}
\rebuttal{In fact, a direct frontend~\cite{bloesch2015robust} is used to estimate the state of a racing drone in~\cite{foehn2022alphapilot}.}
On the contrary, feature-based methods~\cite{mourikis2007multi, leutenegger2015keyframe, qin2018vins} extract points of interest, commonly known as visual features or keypoints, from the raw image pixels.
The camera trajectory is estimated by tracking these points through consecutive images.
\final{High-speed motions make it difficult (e.g., due to motion blur) to track features on many consecutive; consequently, feature-based methods struggle in drone racing scenarios. However, feature-based methods exhibit higher robustness than direct methods to brightness changes. The VIO methods used in~\cite{kaufmann2019beauty, loquercio2019deep} demonstrate that a hybrid frontend, combining the benefits of direct and feature-based methods, is beneficial for drone racing tasks.}

The backend fuses the output of the fronted with the inertial measurements.
Two categories exist in the literature according to how the sensor fusion problem is solved: filtering methods~\cite{mourikis2007multi} and fixed-lag smoothing methods~\cite{leutenegger2015keyframe, qin2018vins}.
Filtering methods are based on an Extended Kalman Filter (EKF).
These methods propagate the system's state using the inertial measurements and fuse the camera measurements in the update step.
The pioneer filter-based VIO algorithm is the Multi-State Constraint Kalman Filter (MSCKF) originally proposed in~\cite{mourikis2007multi}.
Since then, many different versions of MSCKF have been developed~\cite{geneva2020openvins}.
Fixed-lag smoothing methods, also called sliding window estimators,
solve a non-linear optimization problem where the variables to be optimized are a window of the recent robot states. 
The cost function to minimize contains visual, inertial, and past states marginalized residuals.
\rebuttal{Thanks to their favorable trade-off between accuracy and computational cost, filter-based methods have been commonly used in drone racing~\cite{kaufmann2019beauty, loquercio2019deep, foehn2022alphapilot}.}

\subsubsection{Additional sensor modalities in VIO}

\rebuttal{Recently, classical VIO pipelines have been augmented with event cameras~\cite{vidal2018ultimate, sun2021autonomous, chen2023esvio} or drone dynamics~\cite{nisar2019vimo, cioffi2023hdvio, cioffi2023learned}, to improve state estimation at high speed.}

\rebuttal{Low latency, high temporal resolution (in the
order of $\mu s$), and high dynamic range (140 dB compared to 60 dB of standard cameras) are the main properties of event cameras~\cite{gallego2020event}, which make this novel sensor a great complementary sensor to standard cameras.
Including event data in VIO algorithms onboard flying vehicles achieves increased robustness against motion blur as demonstrated in~\cite{vidal2018ultimate, sun2021autonomous, chen2023esvio}. 
Although applications of event cameras in drone racing tasks are yet to be explored, investigating the use of this sensor is a promising research direction to robustify VIO systems for agile flights.}

\rebuttal{The drone dynamics are used to define additional constraints in the estimation process.
The work in~\cite{nisar2019vimo} (VIMO) is the first to integrate error terms related to the drone transitional dynamics in a VIO backend.
VIMO inspired a few works~\cite{ding2021vid, cioffi2023hdvio} which propose an improved noise model of the dynamics~\cite{ding2021vid} and a learned component to account for unmodeled aerodynamics~\cite{cioffi2023hdvio}.
In particular, the results of~\cite{cioffi2023hdvio} show that the learned aerodynamics effects help to improve the VIO estimates at high speeds.}

\rebuttal{The work in~\cite{cioffi2023learned} proposes an odometry algorithm that relies on an IMU as the only sensor modality (no camera is needed), and leverages a learned dynamics component to estimate the state of the racing drone. Consequently, this method does not use a visual frontend.}

\subsubsection{Discussions}

The work in~\cite{delmerico2018benchmark} presents a benchmark comparison between a number of VIO solutions on the EuRoC dataset~\cite{burri2016euroc}.
The EuRoC dataset contains camera and IMU data recorded onboard a drone flying in indoor environments.
The drone moves with average linear and angular velocities up to~\unit[0.9]{m/s} and~\unit[0.75]{rad/s}, respectively.
These values are far below the ones reached in drone racing.
The conclusions of~\cite{delmerico2018benchmark} show that state-of-the-art VIO algorithms provide reliable solutions for estimating the state of the drone at limited speeds.
However, these classical VIO methods cannot provide accurate state estimates for drone racing tasks.
VIO methods accumulate large drift in scenarios characterized by motion blur, low texture, and high dynamic range~\cite{Foehn20rss}.
These scenarios are the norm in drone racing.

To help research VIO algorithms for drone racing tasks, the work in~\cite{delmerico2019UZH} proposes the UZH-FPV Drone Racing Dataset.
This dataset contains images recorded from standard cameras, event camera data, and IMU data recorded onboard a quadrotor flown by a human pilot.
All the flights include visual challenges similar to those in drone racing competitions.

Successful state estimation solutions for drone racing~\cite{kaufmann2018DDR, Foehn20rss} reduce the drift accumulated in VIO by localizing to a prior map of the track.
\rebuttal{In drone racing competitions, a map of the track in the form of gate positions is known beforehand.}
The localization process is based on the detection of the gates. \rebuttal{Performing gate detection is challenging. Often during the race, none of the gates is visible in the camera's field of view. 
Moreover, motion blur makes gate corner detection difficult. For this reason, gate detection and VIO are complementary.}
In~\cite{li2020autonomous}, a gate detector was proposed that uses an RGB camera to identify the gates based on their color. \change{This detector, tailored to the IROS drone racing context~\cite{moon2017iros}, is aimed at extreme computational efficiency, which is particularly important for tiny drones.}
\rebuttal{The method in~\cite{li2020visual} relies on detecting gates and using a model of the drone dynamics to estimate the position of the racing drone.
Differently from~\cite{kaufmann2018DDR,Foehn20rss}, this method does not use a VIO but controls the drone based on a visual-servoing approach.}
All the other gate detection methods in the literature are based on deep learning techniques~\cite{Zhang2020detection}. We review them in Sec.~\ref{Learning-based-Approaches}.
The known gate positions and the detections in the onboard images are used to estimate the relative pose between the camera and the gate using the Perspective-n-Point algorithm (PnP)~\cite{szeliski2022computer}.
This relative pose is used to constrain the VIO estimates and consequently reduce the drift.
There is significant room for innovation on this front, as the VIO-PnP paradigm has existed for several years with little innovation. \change{However, one clear benefit of the VIO-PnP approach is its ability to use a monocular camera setup with a large FOV. While this comes with a lack of scale and higher uncertainty in motion estimation, both can be compensated using inertial sensors and localization with respect to known landmarks (e.g., gates). As evidenced by the rich literature, this makes a monocular setup the preferred solution for autonomous drone racing practitioners. The choice of a monocular sensor is very much in agreement with how human pilots fly: while they have goggles with two monitors, the video stream they receive is from a monocular camera system on the drone.
}
Other approaches used in early drone racing competitions relied on the technique of visual servoing via stereo cameras~\cite{jung2018direct}, but 
\change{relying on a stereo camera pair comes with inherent difficulties. In the presence of motion-blur stereo-matching approaches degrade quickly. Furthermore, drones only allow for a very small baseline and require a wide-angle camera to perceive as much of the surroundings as possible. Both lead to very high depth estimation errors in the stereo setup. The solution proposed in~\cite{jung2018direct}} was found to be sensitive to indoor lighting changes and needed to be hand-tuned for every flight.

Recent works~\cite{wang2017deepvo, wang2020tartanvo, teed2022deep} proposed vision-based odometry algorithms that are learned end-to-end.
Theoretically, these methods could be specialized to drone racing tasks and potentially outperform classical VIO approaches.
However, they are in the early development phase, and how to customize them for the drone racing task is still an open research question. 
In addition, they currently have high computational costs that make them impractical for online state estimation onboard drones. 
We refer the reader to Sec.~\ref{Learning-based-Approaches} for a detailed review of VIO methods based on deep learning.

\subsection{Planning}
\label{sec:traditional_planning}
Once a state estimate $\hat{\bm{x}}$ has been obtained from the perception module, the next step in the classical pipeline is to plan a feasible, time-optimal trajectory ${\tau_{ref} = (\bm{x}_{ref}, \bm{u}_{ref})_k\text{,} \;\; \forall k \in 0 \dots N}$, which respects the physical limits of the platform as well as the constraints imposed by the environment.
This requires predicting the drone's future states such that minimum lap time is reached without crashing.

The planning for drones has matured over the last decade from works mostly verified in simulation to works shown in both controlled lab environments and unknown unstructured environments. 
\change{In the classical pipeline, planning can include up to two distinct planning problems, path planning and trajectory planning.
Path planning tackles the problem of finding a geometrical path between a given start and goal position while passing specified waypoints and avoiding obstacles.
Trajectory planning then uses a found geometric path to either create a collision-free flight corridor\cite{Han2021Fast-racing,spedicato2017minimum}, to find new waypoints for the trajectory to avoid collisions~\cite{richter2016polynomial,penicka22RALmintimeplanning}, to constrain the trajectory to stay close to the found path~\cite{zhou2020robust,zhou2021raptor} or directly finds time allocation for a given path~\cite{pham2018topp,Spasojevic_2020TOPP_quad}.
Therefore, path planning can be seen as a way to select the homotopy class of the collision-free space the drone flies through, while trajectory planning finds the full (or simplified) time-allocated drone state predictions to be followed by the control pipeline (Section~\ref{sec:control}).
However, many works rely solely on trajectory planning as they assume no collision with the environment when a trajectory is found~\cite{penin2018planningvision,foehn2021time, Foehn17rss,Hehn12ar,bousson20134d}.
Other works directly find a collision-free trajectory~\cite{search_planning_lq_mt_Kumar_2017, search_planning_SE3_Kumar_2018,Allen16gnc,Zhiling2020RRTstar_min_jerk} without having a previously found path.
On the other hand, some control approaches~\cite{romero2021model,romero2022replanningRAL} do not need a specified time-allocated trajectory and rely only on the geometrical path for controlling the drone.

In the following text, we first overview the most popular \emph{path planning} approaches for drones that are used for further trajectory planning.
Then, we categorize trajectory planning methods in \emph{polynomial and spline trajectory} planning, \emph{optimization-based} trajectory planning, \emph{search-based} trajectory planning, and \emph{sampling-based} trajectory planning.

\subsubsection{Path planning}
Path planning approaches can be broadly divided into Sampling-based planning and Combinatorial planning~\cite{lavalle2006planning}.
Sampling-based methods do not construct the obstacle space explicitly but rather rely on random sampling of the configuration space together with collision detection.
The most popular variants of the sampling-based methods with numerous modified versions are the Probabilistic Roadmaps~(PRM)~\cite{Kavraki96PRM} and Rapidly-exploring random trees~(RRT)~\cite{Lavalle2000RRT}.
Important variants of these algorithms named  RRT* and PRM*~\cite{Karaman2011OptimalSamplingBased}, can find the optimal path given infinite time. 
The combinatorial planning methods, in contrast to the sampling-based methods, directly represent the obstacle or free space using e.g. polygonal maps or cell decomposition such as grid-based maps.
With the help of a graph representation of the decomposed free space, classical path search algorithms such as A*~\cite{Hart1968Astar} or Dijkstra's algorithm~\cite{dijkstra1959note} can be used to find a path.

Variants of the above path planning approaches are used in many of the methods listed in Sections~\ref{sec:poly_planning}--\ref{sec:sb_planning} to help find trajectories for either fast flight or even drone racing.
The RRT* algorithm is used to find new waypoints for polynomial trajectory planning in~\cite{richter2016polynomial}, and the PRM* is used as a path planning part to guide sampling-based trajectory planning in~\cite{penicka22RALmintimeplanning}.
The sampling-based planning in~\cite{Allen16gnc,Zhiling2020RRTstar_min_jerk} directly performs both path and trajectory planning.
While the trajectory planning objective can be to minimize time duration of a trajectory, the path planning typically tries to find the shortest paths.
Therefore, some methods search for multiple distinct paths to enable the trajectory planning to search over multiple options on how to navigate around obstacles~\cite{penicka22RALmintimeplanning,zhou2020robust,zhou2021raptor}.
Other methods~\cite{Han2021Fast-racing} use search-based algorithms to find an initial path and to create a convex flight corridor for constraining the collision-free trajectory planning.
Similarly, the search-based methods~\cite{search_planning_lq_mt_Kumar_2017, search_planning_SE3_Kumar_2018} use a variant of A* to perform both path and trajectory planning at the same time.
}

\subsubsection{Polynomial and Spline \change{Trajectory Planning}\label{sec:poly_planning}}

The Polynomial and Spline methods leverage the differential flatness property~\cite{Mellinger11icra, Mellinger12ijrr} of quadrotors and represent a trajectory as a continuous-time polynomial or spline.
This property simplifies the full-state trajectory planning to a variant where only four flat outputs need to be planned (typically 3D position and heading). 
By taking their high-order derivatives, these flat outputs can represent a dynamically feasible trajectory with their respective control inputs.
This property is used by many polynomial and spline methods that are nowadays among the most used for general quadrotor flight.

The widely used polynomial trajectories~\cite{Mellinger11icra,Mellinger12ijrr} minimize snap (4\textsuperscript{th} order position derivative) of a trajectory.
Different methods opted for minimizing jerk (3\textsuperscript{rd} order position derivative) for planning a trajectory~\cite{Mueller_minjerk_trajectory}.
However, the trajectories that result from having jerk as the primary objective have been shown to minimize the aggressiveness of the control inputs \cite{Mueller_minjerk_trajectory}, which is fundamentally different from minimizing the lap times, where extremely aggressive trajectories are generally required.
Richter et al.~\cite{richter2016polynomial}, therefore, extended the objective by jointly optimizing both the snap of a trajectory and the total time through a user-specified penalty on time.
Recently, Han~\cite{Han2021Fast-racing} proposed a polynomial-based trajectory planning method for drone racing. 
It jointly optimizes control effort and regularized time and penalizes the dynamic feasibility and collisions.

Because of their numerical stability, other methods use B-splines to represent trajectories~\cite{zhou2020robust,zhou2021raptor} instead of high-order polynomial representations that are numerically sensitive.
These methods jointly optimize different objectives, simultaneously smoothness, dynamic feasibility, collision avoidance, safety~\cite{zhou2021raptor} and vision-based target tracking~\cite{penin2018planningvision}.
\final{Recently, the authors of~\cite{qin2023time} proposed a polynomial trajectory representation based on the work of~\cite{WANG2022GCOPTER} and use it to plan time-optimal trajectories through gates of arbitrary shapes for drone racing, achieving close-to-time-optimal results while being more computationally efficient than \cite{foehn2021time}.}

Although both polynomial and spline trajectories are widely used due to their computational efficiency, polynomial-based trajectories (and their derivatives) are smooth by definition. Therefore, only smooth control inputs can be sampled from them.
\final{%
For this reason, the traditional polynomial planning~\cite{Mellinger11icra} with a finite number of coefficients and one polynomial segment between every two waypoints (gates) cannot represent true time-optimal trajectories~\cite{foehn2021time}.
Yet, direct collocation methods~\cite{fork2023euclidean} that rely on polynomials to approximate the input and state dynamics can achieve nearly optimal performance. 
This is mainly due to a larger number of polynomial segments between the waypoints in collocation methods, joint optimization of both polynomial coefficients and collocation points, and due to the approximation of the entire dynamics by polynomials. 
This allows to keep the acceleration at the possible maximum at all times similar to the optimization-based shooting method~\cite{foehn2021time}.
Therefore, while the classical polynomial and spline methods can be considered optimization-based, they only optimize coefficients of a single polynomial between every two waypoints to describe quadrotor position and heading, leveraging the differential flatness property~\cite{Mellinger11icra, Mellinger12ijrr}.

}

\subsubsection{Optimization-based \change{Trajectory Planning}}

Optimization-based trajectory planning enables us to independently select the optimal sequence of states and inputs at every time step, which inherently considers time minimization while complying with quadrotor dynamics and input constraints. 
Optimization-based approaches have been extensively considered in the literature, ranging from exploiting point-mass models~\cite{Foehn17rss}, simplified quadrotor models~\cite{Hehn12ar, Loock13ecc}, and full-state quadrotor models~\cite{foehn2021time, spedicato2017minimum}.

Time-optimality of a trajectory could also be accomplished by using a specific path parameterization that maximizes velocity over a given path\cite{pham2018topp}.
This method was shown for quadrotors in \cite{Spasojevic_2020TOPP_quad} for minimizing time of flight considering both translational and rotational quadrotor dynamics.
However, the method only creates a velocity profile over a given path which is not further optimized.

Apart from time optimality, complying with intermediate waypoint constraints is another requirement for path planning in autonomous drone racing. 
A common practice of solving a trajectory optimization problem with waypoint constraints is allocating waypoints to specific time steps and minimizing the spatial distance between these waypoints and the position at the corresponding allocated time steps on the reference trajectory (e.g. \cite{jorris2009three, bousson20134d}). 
The time allocation of the waypoints is, however, non-trivial and difficult to determine. 
This is tackled in~\cite{spedicato2017minimum}, but the work uses body rates and collective thrust as control inputs and does not represent realistic actuator saturation.
Recent work~\cite{foehn2021time} introduces a complementary progress constraints (CPC) approach, which considers true actuator saturation, uses single rotor thrusts as control inputs, and exploits quaternions to create full, singularity-free representations of the orientation space with consistent linearization characteristics.
While the above methods create time-optimal trajectories passing through given gates, they are computationally costly and hence intractable in real-time.

\subsubsection{Search-based \change{Trajectory Planning}}
Search-based planning methods~\cite{search_planning_lq_mt_Kumar_2017, search_planning_SE3_Kumar_2018} rely on discretized state and time spaces.
They solve the trajectory planning through graph search algorithms such as A*.
The search graph is built using minimum-time motion primitives with discretized velocity, acceleration, or jerk input.
The algorithms then use trajectories of a simpler model, e.g. with velocity input, as heuristics for the search with a more complex model.
Search-based planning methods can optimize the flight time up to discretization, but they suffer from the curse of dimensionality which renders them too computationally demanding for a complex quadrotor model.
Furthermore, the employed per-axis dynamic limits (velocity, acceleration, jerk) do not represent the true quadrotor model, further decreasesing the quality of found plans.
Finally, although searching for minimum time trajectories, the methods are currently limited to planning between two states which is not suitable for multi-waypoint drone racing.

\subsubsection{Sampling-based \change{Trajectory Planning}\label{sec:sb_planning}}
Sampling-based methods like RRT*~\cite{webb2013KinodynamicRRTstar} can be used for planning trajectories for linearized quadrotor models.
Several time-minimizing approaches~\cite{Foehn20rss, Allen16gnc} use a point-mass model for high-level time-optimal trajectory planning. 
In~\cite{Allen16gnc}, an additional trajectory smoothing step is performed where the generated trajectory is connected with high-order polynomials by leveraging the differential flatness property of the quadrotor. 
\change{Authors of~\cite{ichter2020perception} use sampling-based approach with massive GPU parallelization and a 6D double integrator system of UAV with additional single integrator yaw dynamics.}
However, these point-mass approaches need to relax the single actuator constraints and instead limit the per-axis acceleration, which results in trajectories that are conservative and sub-optimal given a minimum time objective.
In~\cite{Zhiling2020RRTstar_min_jerk}, the authors use minimum-jerk motion primitives for connecting randomly sampled states inside RRT* to plan a collision-free trajectory. 
Since the authors use polynomials, this approach can only generate smooth control inputs, meaning that they cannot rapidly switch from full thrust to zero thrust if required.

The first method for planning minimum-time trajectories in a cluttered environment for the full quadrotor model was proposed in~\cite{penicka22RALmintimeplanning}.
It uses a hierarchical sampling-based approach with an incrementally more complex quadrotor model to guide the sampling.
The authors showed that the method outperforms both polynomial and search-based methods in minimizing trajectory time.
Yet, the method is offline and intractable in real-time.
Most recently, the authors of~\cite{romero2022replanningRAL} proposed an online replanning approach that plans minimum-time trajectories for a point-mass model.
The paths of replanned trajectories are then consequently used by Model Predictive Contouring Control~\cite{romero2021model} with a full quadrotor model to maximize the progress along the path.
This method is capable of outperforming other classical approaches due to the replanning capability and progress maximization with a full quadrotor model.

\subsubsection{Discussion}

A planned trajectory can be understood as an intermediate representation that, given information about the robot's dynamics and the environment, helps guide the platform through the race track and ultimately perform the task at hand. 
One might argue if this intermediate representation is needed at all, since ultimately, what we are looking for is a policy that maps sensor information and current environment knowledge to the actuation space.
This is generally achieved with learning-based approaches, discussed in Section \ref{Learning-based-Approaches}, which bypass the planning stage and directly convert sensor observations to actuation commands~\cite{kaufmann2020deep,song2021autonomous,penicka2022learning}.

One of the biggest benefits of explicit planning is  \emph{modularity}.
This means that the developed algorithms can be used off-the-shelf for different drone tasks outside racing, such as search and rescue, which is not the case for single-purpose learned approaches.
However, explicit planning suffers from the disconnection (or an open loop) between the planning and the deployment stage.
Unexpected deviations from the plan, be it in the time domain (like unmodeled system delays) or in the state-space domain (like state estimation drifts or jumps in the VIO pipeline), can lead to compound errors and ultimately, a complete system failure.

This can be tackled with more complex control approaches that do some part of the replanning online~\cite{romero2022replanningRAL}.

\subsection{Control\label{sec:control}}
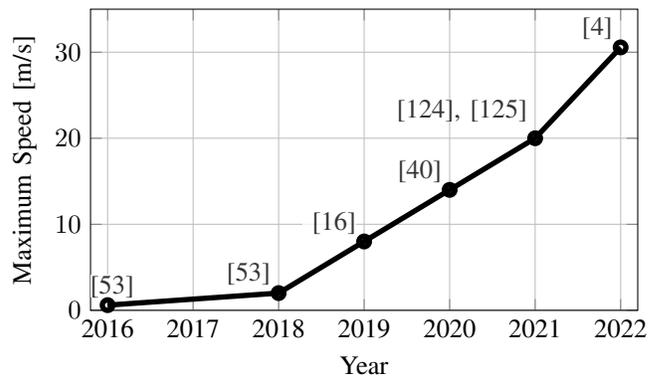
\begin{figure}[t]
    \centering
    \begin{tikzpicture}[]

\begin{axis}[%
width=\columnwidth,
height=2.2in,
at={(0.758in,0.481in)},
xmin=2015.8,
xmax=2022.2,
xlabel={Year},
ymin=0,
ymax=35,
ylabel={Maximum Speed [m/s]},
axis background/.style={fill=white},
xmajorgrids,
ymajorgrids,
xtick={2016, 2017, 2018, 2019, 2020, 2021, 2022},
xticklabels = {2016, 2017, 2018, 2019, 2020, 2021, 2022}
]
\node at (axis cs:2016, 0.6) [anchor=south,outer sep=3pt,, inner sep=0pt,, fill=white, fill opacity=0.8] {\cite{perez2021onboard}};
\node at (axis cs:2018, 2.0) [anchor=south east, outer sep=3pt,, inner sep=0pt,, fill=white, fill opacity=0.8] {\cite{perez2021onboard}};
\node at (axis cs:2019, 8.0) [anchor=south east, outer sep=3pt,, inner sep=0pt,, fill=white, fill opacity=0.8] {\cite{foehn2022alphapilot}};
\node at (axis cs:2020, 14.0) [anchor=south east, outer sep=3pt,, inner sep=0pt,, fill=white, fill opacity=0.8] {\cite{torrente2021datadriven}};
\node at (axis cs:2021,21) [anchor=south east, outer sep=3pt,, inner sep=0pt,, fill=white, fill opacity=0.8] {\cite{hanover2021performance, sun2021comparative} };
\node at (axis cs:2022,30.56) [anchor=south east, outer sep=3pt,, inner sep=0pt,, fill=white, fill opacity=0.8] {\cite{Song23Reaching} };
\addplot [color=black,
        line width=2.0pt,
        mark=o,
        mark options={solid, black},
        forget plot,
        ] table [%
     ] {
x       y       
2016	0.6   
2018	2    
2019	8   
2020	14  
2021	20  
2022	30.56 
};
\end{axis}

\end{tikzpicture}%
    \caption{Top speeds demonstrated on autonomous drones over time from both literature and competition data.}
    \label{fig:topSpeeds}
\end{figure}

Over the last decade, significant advancements have been made in agile multicopter control.
Every year, increasing top speeds are demonstrated in the literature as shown in Figure~\ref{fig:topSpeeds}.

Controllers must be able to make real-time decisions in the face of poor sensor information and model mismatch. 
Control inputs, $u(t)$, can come in a variety of modalities for quadrotor control, such as velocity and heading, body rates and collective thrust, or direct rotor thrust commands~\cite{kaufmann2022benchmark}.
Typically, a high-level controller computes a desired virtual input such as body rates and collective thrust, which is then passed down to a low-level flight controller that directly controls the individual rotors on the multicopter.

Commonly used open source controllers such as PixHawk\footnote{https://pixhawk.org/} 
or BetaFlight\footnote{https://github.com/betaflight/betaflight} are widely available to the drone racing community. 
BetaFlight is the most commonly used low-level controller for agile drone flight and has been widely adopted by the First Person View (FPV) racing community.

In the following sections, we provide an overview of successful approaches to achieving high speeds in both simulation and real-world applications.
We sort the approaches into \textit{model-based control} and \textit{coupled perception and control}.

\subsubsection{Model-Based Control}
In model-based control, an explicit model of the dynamic system is used to calculate control commands that satisfy a given objective such as minimizing time or tracking error. 
Models enable the prediction of future states of the drone and provide information about the system's stability properties.
In \cite{Lee2011Geo}, Geometric Tracking control is introduced on the Special Euclidean group SE(3) and completely avoids singularities commonly associated with Euler angle formulations on SO(3).
This nonlinear controller showed the ability to execute acrobatic maneuvers in simulation and was the first to demonstrate recovery from an inverted initial attitude. 
The dynamic model of a quadrotor is shown to be differentially flat when choosing its position and heading as flat outputs in \cite{Mellinger12ijrr}.
In this work, many agile maneuvers are performed onboard real drones with speeds up to \unit[2.6]{m/s}.

The previous work is extended in \cite{faessler2017differential}, proving that the dynamics model of a quadrotor subject to linear rotor drag is also differentially flat.
The inclusion of the aerodynamic model within the nonlinear controller led to demonstrated flight speeds up to \unit[4.0]{m/s} while reducing tracking error by 50\% onboard a real drone.

The differential flatness method is further extended in~\cite{tal2020accurate} by cascading an Incremental Nonlinear Dynamic Inversion (INDI) controller with the differential flatness controller described in \cite{Mellinger12ijrr} but neglects the aerodynamic model addition from \cite{faessler2017differential}.
The INDI controller is designed to track the angular acceleration commands $\dot{\Omega}$ from the given reference trajectory.
Top speeds of nearly \unit[13]{m/s} and accelerations over \unit{2}{g} are demonstrated onboard a real quadrotor. 
The controller shows robustness against large aerodynamic disturbances in part due to the INDI controller.

An investigation of the performance of nonlinear model predictive control (NMPC) against differential flatness methods is available in~\cite{sun2021comparative}.
Cascaded controllers of INDI-NMPC and INDI-differential flatness are shown to track aggressive racing trajectories which achieve speeds of around \unit{20}{m/s} and accelerations of over \unit{4}{g}. 
While differential flatness methods are computationally efficient controllers and relatively easy to implement, they are outperformed on racing tasks by NMPC.

An excellent overview of MPC methods applied to micro aerial vehicles can be found in~\cite{nguyen2021model}.
Because quadrotors are highly nonlinear systems, nonlinear MPC is often used as the tool of choice for agile maneuvers.
The debate of linear versus nonlinear MPC is thoroughly discussed in \cite{kamel2017linear}.
Model Predictive Path Integral (MPPI) control is a sampling-based optimal control method that has found excellent success on the AutoRally project, a 1/5th scale ground vehicle designed to drive as fast as possible on loose dirt surfaces \cite{williams2017model,goldfain2019autorally}.
An introduction to MPPI can be found in \url{https://autorally.github.io/}.
The MPPI approach can be used on agile quadrotors to navigate complex forest environments, however, analysis was only performed in simulation \cite{williams2017model}.
Most of the successful demonstrations of MPPI come from ground robots \cite{williams2017model,goldfain2019autorally}.
Because MPPI is a sampling based algorithm, scaling to higher-dimension state spaces of quadrotors can lead to performance issues as shown in \cite{hanover2021performance}.

Nonlinear MPC methods are also used in~\cite{torrente2021datadriven} where a nominal quadrotor model is augmented with a data-driven model composed of Gaussian Processes and used directly within the MPC formulation.
The authors found that the Gaussian-Process model could capture highly nonlinear aerodynamic behavior which is difficult to model in practice as described in Sec.~\ref{sec:modeling}.
The additional terms introduced by the Gaussian-Process added computational overhead to the MPC solve times, but it was still able to run onboard a Jetson TX2 computer.

Similar to~\cite{tal2020accurate}, authors in~\cite{hanover2021performance} question whether or not it is necessary to explicitly model the additional aerodynamic terms from~\cite{torrente2021datadriven} due to the added computational and modeling complexity.
Instead, they propose to learn residual model dynamics online using a cascaded adaptive nonlinear model predictive control architecture.
Aggressive flight approaching \unit{20}{m/s} and over \unit{4}{g} acceleration is demonstrated on real racing quadrotors.
Additionally, completely unknown payloads can be introduced to the system, with minimal degradation in tracking performance.
The adaptive inner loop controller added minimal computational overhead and improved tracking performance over the Gaussian Process MPC by 70\% on a series of high-speed flights of a racing quadrotor~\cite{hanover2021performance,torrente2021datadriven}.  

Contouring control methods can deal with competing optimization goals such as trajectory tracking accuracy and minimum flight times~\cite{Lam2010MPCC}.
These methods minimize a cost function which makes trade-offs between these competing objectives.
In \cite{Liniger_2014}, Nonlinear Model Predictive Contouring Control (MPCC) is applied to control small model racecars.
MPCC was then extended to agile quadrotor flight in ~\cite{romero2021model}.
Although the velocities achieved by the MPCC controller were lower than that of \cite{hanover2021performance,sun2021comparative}, the lap times for the same race track were actually lower due to the ability of the controller to find a new time-allocation that takes into account the current state of the platform at every timestep.
The work is further extended to solve the time-allocation problem online, and to re-plan online \cite{romero2022replanningRAL} while also controlling near the limit of the flight system. 
Similar work uses tunneling constraints in the MPCC formulation in \cite{arrizabalaga2021towards}, 

\change{
\subsubsection{Perception Awareness}
Other methods that lie in the intersection of perception, planning, and control include a perception objective in the cost function that helps improve the visibility of an objective or the quality of the state-estimation pipeline. The methods are called \emph{perception aware}, and the first methods were proposed in \cite{costante2016perception, falanga2017aggressive, penin2017vision} 
This is integral to the drone-racing problem because, to navigate a challenging race course, the gates that define the course layout must be kept in view of the onboard cameras \change{as much as possible}.
Additionally, coupling the perception with the planning and/or control problem can alleviate issues in state estimation because the racing gates are usually feature-rich.
Therefore, the use of perception-related objectives in both planning and control pipelines is commonplace \cite{penin2017vision, ichter2020perception, zhou2021raptor, tordesillas2023deeppanther, Spasojevic_2020TOPP_quad}.
For example, in \cite{penin2017vision, Spasojevic_2020TOPP_quad} the authors tackle the problem of minimizing the time required by a quadrotor to execute a given path, while maintaining a given set of landmarks within the field of view of its on-board camera.
Or in \cite{tordesillas2022panther}, where the authors include a perception-aware term in the cost function to maximize the visibility of the closest dynamic obstacle, in order to readily plan a path that avoids it.
These methods are called \emph{perception-aware} \cite{falanga2018pampc}, and in the following, we highlight their core characteristics.
}

The goal is as follows: navigate a trajectory with low tracking error while keeping a point of interest in view while minimizing motion blur for maximum feature detection and tracking. 
The first instance applied to agile quadrotors was PAMPC introduced in \cite{falanga2018pampc}.
In this work, a nonlinear program is optimized using a sequential quadratic programming approximation in real time. 
The cost function contains both vehicle dynamic terms as well as perception awareness terms such as keeping an area of interest in the center of the camera frame.

This technique is applied to the drone racing problem in \cite{lee2020aggressive}, where an MPPI controller is designed with a Deep Optical Flow (DOF) component that predicts the movement of relevant pixels (i.e. gates).
The perception constraints are introduced into a nonlinear optimization problem and deployed in a drone-racing simulator.
The approach was not demonstrated onboard real hardware.
In \cite{greeff2020perception}, a perception-aware MPC based on Differential Flatness was used to ensure that a minimum number of features are tracked between control updates and thus guarantee localization.
To achieve this, a Perception Chance Constraint within the MPC formulation is introduced to ensure that at least $n$ number of landmarks are within the field-of-view of the camera at all times with some bounded probability.

\subsubsection{Discussion}
The performance of model-based controllers degrades when the model they operate on is inaccurate~\cite{hanover2021performance}. 
For drones, defining a good enough model is an arduous process due to highly complex aerodynamic forces, which can be difficult to capture accurately within a real-time capable model. 
In addition, the tuning process of many model-based controllers can be arduous, and requires a high level of domain expertise to achieve satisfactory performance\change{.}

In any optimal control problem, a cost function that the user wants to optimize must be defined.
Traditionally, convenient mathematical functions leveraging convex costs are used because these functions are easy to optimize and there is a large toolchain available for optimizing such problems such as Acados~\cite{verschueren2022acados}, CVXGEN~\cite{mattingley2012cvxgen}, HPIPM~\cite{frison2020hpipm}, or Mosek~\cite{aps2019mosek}.
In many drone racing papers, the optimal control problem is formulated as follows:
 \begin{gather}
     \min\limits_{\bm{u}} \bm{x}_N^T \bm{Q} \bm{x}_N + \sum_{k=0}^{N-1} \bm{x}_k^T \bm{Q} \bm{x}_k + \bm{u}_k^T \bm{R} \bm{u}_k \text{ ,}\\
     \st \quad \bm{x}_{k+1} = \bm{f}_{RK4}(\bm{x}_k, \bm{u}_k, \delta t)\nonumber \text{ ,}\\
     \begin{aligned}
     \bm{x}_0 &= \bm{x}_{init}\nonumber \text{ ,} &
     \bm{u}_{min} &\leq \bm{u}_k \leq \bm{u}_{max}\nonumber \text{ ,}
     \end{aligned}
 \end{gather}
where the state is given by $x_k$, the control input is given by $u_k$, the state cost matrix is given by $Q$, and the control cost matrix is given by $R$.
The optimization problem is constrained by the dynamics of the system given by $f(x_k, u_k, \delta t)$ where $\delta t$ is a finite time step.
The nonlinear dynamics are typically propagated forward using an integrator such as 4th order Runge-Kutta, $RK4$.
Additionally, the problem is subject to the thrust limits of the platform. $u_{min}$ and $u_{max}$, and some initial condition of the system $x_0$.
In this formulation, a reference position and control are provided by a high-level planner and the goal of the controller is to track the given reference, but this objective is ill-defined for the drone racing problem:
in drone racing, we wish to complete the track in as little time as possible; therefore, our objective can be better formulated as follows: \vspace*{-6pt}
\rebuttal{
 \begin{gather}
     \min\limits_{\bm{u}} \sum_{k=0}^{T} \delta t \text{ ,}\\
    \begin{aligned}
    \st \quad
    &\bm{x}_{k+1} = \bm{f}_{RK4}(\bm{x}_k, \bm{u}_k, \delta t)\nonumber \\
    &\bm{x}_0 = \bm{x}_{init},\quad
     \bm{x} \in \mathcal{X}, \quad \bm{u} \in \mathcal{U}\\
     \end{aligned}
 \end{gather}
}
where $T$ is the number of discrete time steps it takes to complete the race, \change{and the set $\mathcal{U}$ contains the input constraints (e.g., single-rotor thrust constraints). The set $\mathcal{X}$ encodes all state constraints, from possible limits in the state itself (e.g., attitude or velocity constraints), to more complex constraints such as the fact that the drone has to pass through a set of gates in a pre-determined order without colliding.}
This approach requires a time-horizon that predicts all the way until the end of the task which is intractable to optimize online.

Reinforcement learning (RL) methods~\cite{song2021autonomous,kaufmann2022benchmark} can optimize a proxy of this cost function, however do so in an offline fashion, requiring large amounts of training experience to approximate the value function.
RL methods do not necessarily depend on a high-level planner to provide a reference to track.
We will discuss some recent approaches using reinforcement learning methods in the following section.

\section{Learning-based Approaches}\label{Learning-based-Approaches}
In this section, we present various learning-based approaches for drone racing.
These approaches replace the planner, controller, and/or perception stack with a neural network.
Learning-based methods have gained significant traction in the last few years, given their ability to cope with both high-dimensional (e.g. images) or low-dimensional (e.g. states) input data, their representation power, and the ease of developing and deploying them on hardware.

\change{The big advantage of these methods is that they require less computational effort than traditional methods, possibly enabling
low-latency re-planning and control.
In addition, they are much more robust to system latencies and sensor noise, which can be easily accounted for by identifying them on physical drones and then
adding them to the training environments \cite{kaufmann2022benchmark}.
However, the major limitation of these methods is their sample complexity.
There are currently two possibilities for data gathering. 
The first, mostly popular in the initial stages of learning-based robotics~\cite{giusti2015machine,loquercio2018dronet,kaufmann2018DDR,kaufmann2019beauty, gandhi2017learning} is to collect data in the real world.
The data is then annotated by a human or an automated process, and used for training.
The second, much more popular in recent years and currently achieving the best results, consists of using simulation for collecting training data~\cite{loquercio2019deep,Loquercio21FlightWild,sadeghicad,kaufmann2020deep,kaufmann2022benchmark}.
However, significant simulation engineering is required to enable generalization if the training data comes from a simulator. Conversely, generalization is easier if data come from the real world, but the data collection process is very slow, tedious, and expensive.
}

Surveys covering existing methods for learning-based flight already exist~\cite{Lee2021FlyingFA, pham2022deep}.
In contrast to them, we cover the most recent advances and give a broader discussion on the comparison between learning-based and traditional methods for drone racing.

\subsection{Learned Perception}
\begin{figure}[h]
    \centering
    \vspace*{-12pt}
    \begin{tikzpicture}[>=latex, font=\footnotesize]
    \tikzstyle{tb}=[minimum width = 1.4cm, text width = 1.2cm, minimum height = 0.4cm, align = center, draw, inner sep = 1pt, inner xsep=3pt];
    \newcommand\dx{0.4cm};
    \newcommand\dy{0.05cm};
    \newcommand\shift{0.05cm};
     \node [tb] (a) {Sensors};
     \node [tb, draw = none, above = \dy of a] (a2) {\emph{Hardware}};
     \node [tb, draw=none, right = \dx of a] (b) {};
     \node [tb, draw = none, above = \dy of b] (b2) {};
     \node [tb, right = \dx of b] (c) {Planning};
     \node [tb, right = \dx of c] (d) {Control};
     \node [tb, right = \dx of d] (e) {Drone};
     \node [tb, draw = none, above = \dy of d] (d2) {\emph{Software}};
     \node [tb, draw = none, above = \dy of e] (e2) {};
     \draw [->] (a) -- (b);
     \draw [->] (b) -- (c);
     \draw [->] (c) -- (d);
     \draw [->] (d) -- (e);
     \draw (b2.north west) rectangle (b.south east) node [midway] {\includegraphics[width = 1.2cm]{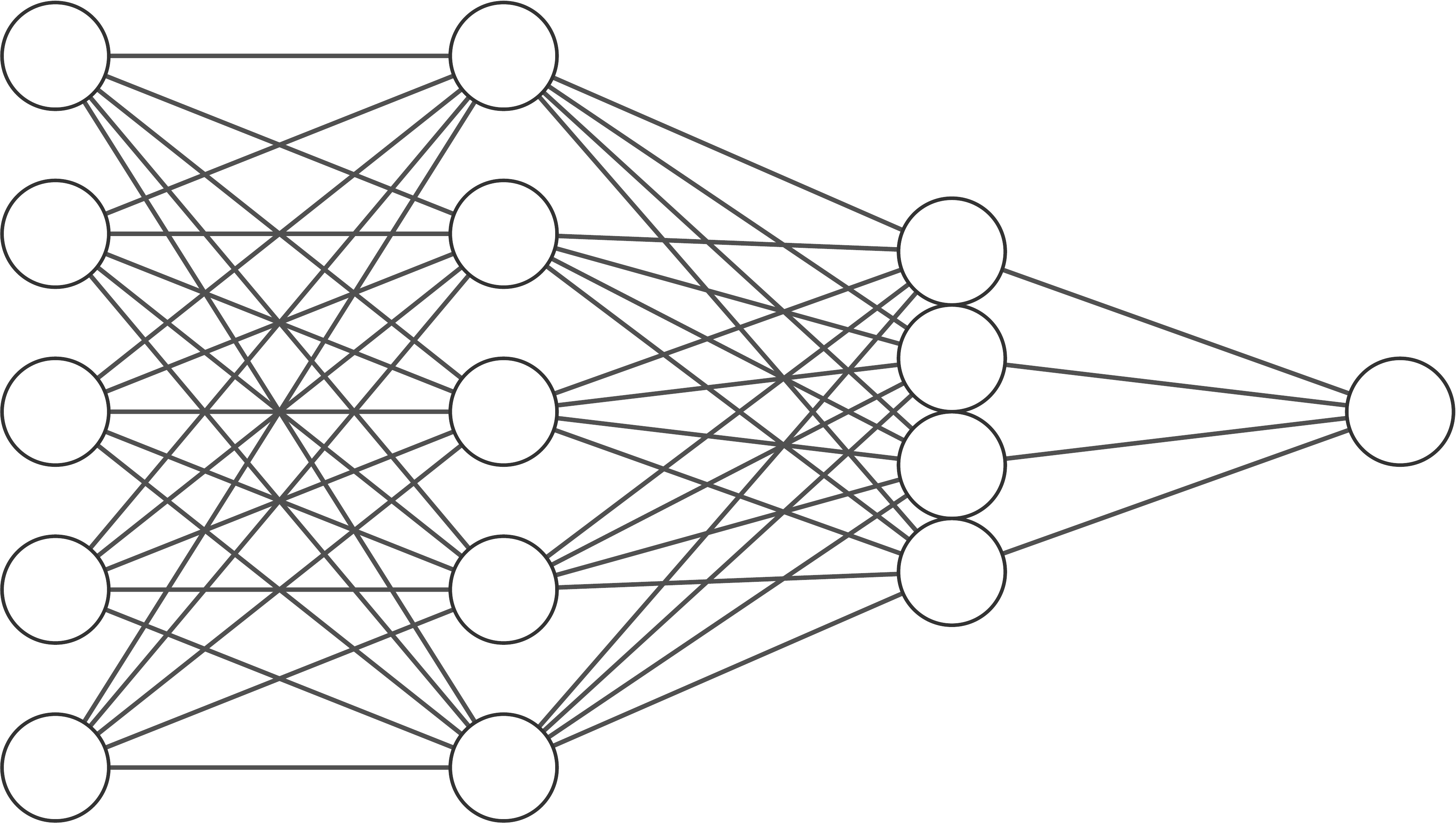}} ;
     \begin{pgfonlayer}{bg}    %
         \draw [draw=none, fill=red!10!white] ([xshift = -\shift, yshift=\shift]b2.north west) rectangle ([xshift=\shift, yshift=-\shift]d.south east);
         \draw [draw = none, fill=blue!10!white] ([xshift = -\shift, yshift=\shift]a2.north west) -| ([xshift=\shift, yshift=-2*\shift]a.south east) -| ([xshift=-\shift, yshift=\shift]e2.north west) -| ([xshift=\shift, yshift=-4*\shift]e.south east) -| ([xshift=\shift, yshift=-4*\shift]a.south east) -| cycle;
    \end{pgfonlayer}
    \end{tikzpicture}
    \vspace*{-8pt}
    \caption{Architecture 2: Learned Perception}
    \label{fig:LearnedPerception}
    \vspace*{-6pt}
\end{figure}
For learned perception modules, the goal of the network is to use images from an RGB, depth, or event camera to detect landmarks within the environment and output useful representations such as waypoints, or the location of gates on the track.
A depiction of this architecture can be seen in Fig.~\ref{fig:LearnedPerception}.
An overview of deep learning methods for vision-based navigation specific to drone racing can be found in~\cite{pham2022deep}.

In~\cite{kaufmann2019beauty}, a dataset of images is collected from a forward-facing camera mounted on a drone labeled with the relative position to the closest gate.
This dataset is used to train a network that predicts from an image both the next gate location and its uncertainty.
Predictions are then fused with a visual-inertial odometry system in an Extended Kalman Filter (EKF) to predict the position of the drone on the track.
Similarly in \cite{foehn2022alphapilot}, a Convolutional Neural Network (CNN) is used to detect gate corners in the AlphaPilot challenge.
Once the gate corners are detected, classical computer vision algorithms like PnP can be used to find the coordinates of the gate in the camera frame.
Using an EKF, the gate corner locations can be fused with a traditional VIO pipeline to improve the estimates of the drone's location and orientation~\cite{foehn2022alphapilot}.

Oftentimes, perception networks consume precious resources onboard computationally limited drones.
To minimize the network processing time, \cite{Zhang2020detection, Cabrera2019gate}  proposed optimized architectures for gate detection on real-world data.
A similar optimization went into ``GateNet"~\cite{pham2021gatenet} a CNN to detect gate center locations, distance, and orientation relative to the drone.
The same authors developed a follow-up work denoted as ``Pencil-Net" to do gate detection using a lightweight CNN in \cite{pham2022pencilnet}.
Most learning-based perception networks can suffer from poor generalization when deployed in environments that were not included in the training data.%
To reduce deployment sensitivity to lighting conditions or background content, virtual gates can be added to real-world backgrounds ~\cite{morales2020image}.

Up until recently, RGB and depth cameras were used exclusively in the drone racing task, however, these sensor modalities can be sensitive to changes in the environment such as illumination changes.
To overcome this, \cite{andersen2022event}~proposed using event cameras coupled with a sparse CNN, recurrent modules, and a You Only Look Once (YOLO) object detector to detect gates.
The use of event cameras overcomes potential issues with motion blur from the rapid movement of the drone and is a promising path forward for high-speed navigation.

Overall, deep learning methods for gate detection are the de-facto standard in all drone racing systems.
However, such gate detectors are always coupled with traditional visual-inertial odometry systems which explicitly estimate the metric state of the drone.
These approaches are discussed in Sec.~\ref{ClassicalPipeline}.
It is interesting to notice that learning-based odometry systems, such as~\cite{wang2017deepvo, wang2020tartanvo, teed2022deep} have not yet replaced traditional methods.
This is particularly surprising since deep visual odometry systems can specialize to a particular environment, which can be useful for drone racing since the race track is fixed and known in advance.
A disadvantage of these methods is the high computational cost that makes them impractical for online applications. 
However, research in end-to-end visual odometry is moving forward at a fast pace~\cite{teed2022deep}.
\rebuttal{Recently, works proposing end-to-end VIO systems for drones have been published~\cite{sanket2021prgflow, xu2021cnn, xu2022cuahn}. The work in~\cite{sanket2021prgflow} proposes to learn global optical flow which is then loosely fused with an IMU for full 6-DoF relative pose estimation.
The method in~\cite{xu2021cnn} and its extension~\cite{xu2022cuahn} proposes a CNN-based ego-motion estimator for fast flights. 
The performance of this method in the UZH-FPV dataset shows that although end-to-end VIO methods are a promising solution for agile flights, they are not yet mature for drone racing.}
We foresee that in the near future, researchers will be able to apply these methods to the drone racing task.

\vspace*{-6pt}
\subsection{Learned Planning \& Perception}
\begin{figure}[H]
    \centering
    \vspace*{-9pt}
    \begin{tikzpicture}[>=latex, font=\footnotesize]
    \tikzstyle{tb}=[minimum width = 1.4cm, text width = 1.2cm, minimum height = 0.4cm, align = center, draw, inner sep = 1pt, inner xsep=3pt];
    \newcommand\dx{0.4cm};
    \newcommand\dy{0.05cm};
    \newcommand\shift{0.05cm};
     \node [tb] (a) {Sensors};
     \node [tb, draw = none, above = \dy of a] (a2) {\emph{Hardware}};
     \node [tb, draw = none, right = \dx of a] (b) {};
     \node [tb, draw = none, above = \dy of b] (b2) {};
     \node [tb, draw = none, right = \dx of b] (c) {};
     \node [tb, draw, right = \dx of c] (d) {Control};
     \node [tb, right = \dx of d] (e) {Drone};
     \node [tb, draw = none, above = \dy of d] (d2) {\emph{Software}};
     \node [tb, draw = none, above = \dy of e] (e2) {};
     \draw [->] (a) -- (b);
     \draw [->] (c) -- (d);
     \draw [->] (d) -- (e);
     \draw (b2.north west) rectangle (c.south east) node [midway] {\includegraphics[width = 2.4cm]{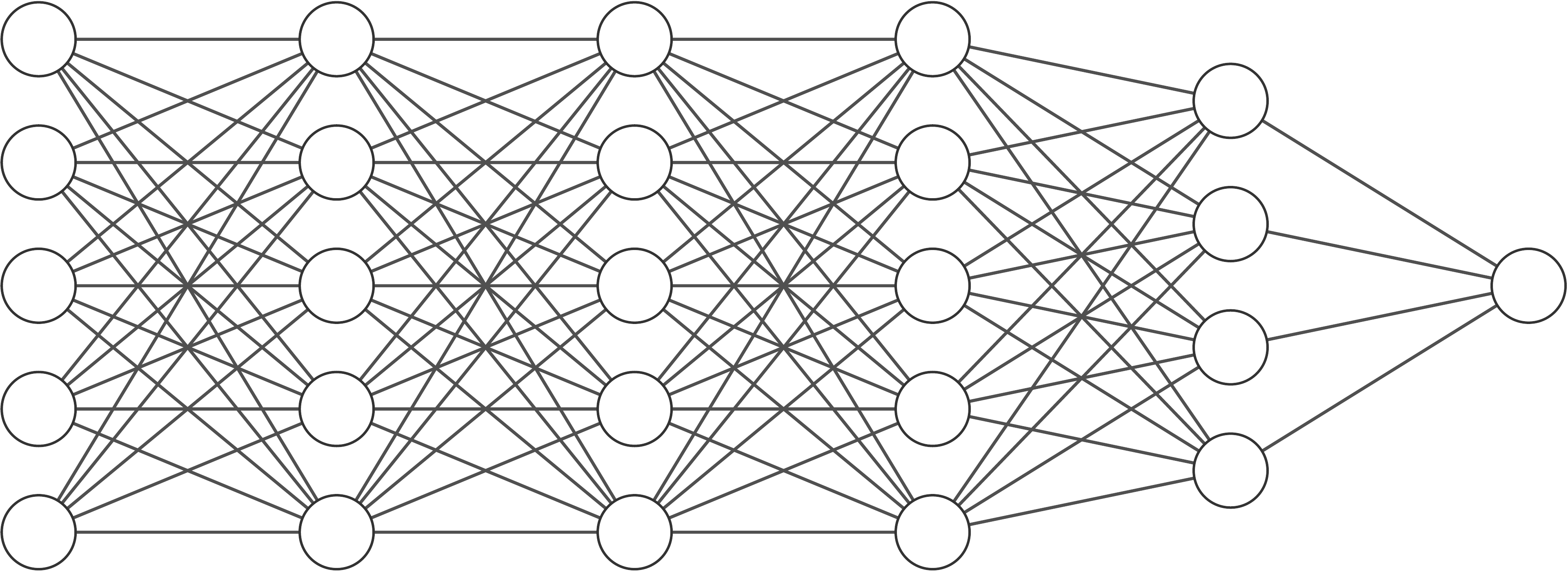}} ;
     \begin{pgfonlayer}{bg}    %
         \draw [draw=none, fill=red!10!white] ([xshift = -\shift, yshift=\shift]b2.north west) rectangle ([xshift=\shift, yshift=-\shift]d.south east);
         \draw [draw = none, fill=blue!10!white] ([xshift = -\shift, yshift=\shift]a2.north west) -| ([xshift=\shift, yshift=-2*\shift]a.south east) -| ([xshift=-\shift, yshift=\shift]e2.north west) -| ([xshift=\shift, yshift=-4*\shift]e.south east) -| ([xshift=\shift, yshift=-4*\shift]a.south east) -| cycle;
    \end{pgfonlayer}
    \end{tikzpicture}
    \vspace*{-8pt}
    \caption{Architecture 3: Learned Planning and Perception}
    \label{fig:LearnedPlanningAndPerception}
    \vspace*{-6pt}
\end{figure}

A tightly-coupled planning and perception stack (Figure \ref{fig:LearnedPlanningAndPerception}) is a very attractive algorithmic perspective.
First, it greatly simplifies the perception task: an explicit notion of a map or globally-consistent metric state is not required.
Second, it largely reduces computational costs, both in the pre-training and evaluation stages.
Finally, it can leverage large amounts of data, collected either in simulation or the real world, to become robust against noise in perception or dynamics.
\change{Yet, an interesting observation is that these methods still work best when coupled with an explicit estimator of the metric state~\cite{kaufmann23champion}.}
In contrast to traditional methods, a locally consistent odometry system is sufficient~\cite{loquercio2019deep,Loquercio21FlightWild,kaufmann2018DDR}, waving away the complexities of full-slam methods (e.g. loop-closure).

In~\cite{kaufmann2018DDR}, a coupled perception and planning stack for drone racing is trained using real-world flight demonstrations.
While good performance is indicated on the racing task as well as robustness against drift in state estimation, the method requires re-training for each new environment.
Therefore, in the follow-up work~\cite{loquercio2019deep}, data generated entirely in simulation is used to train the perception-planning stack, waiving the labor and time-consuming requirement of data collection in the real world.
A similar pipeline was used for high-speed autonomous flight through complex environments in~\cite{Loquercio21FlightWild}, which proposes to train a neural network in simulation to map noisy sensory observations to collision-free trajectories directly.
\final{This approach was later extended to nano-quadcopters~\cite{lamberti2024sim}, which won the authors the first position in the \emph{IMAV 2022 Nanocopter AI Challenge}.}
\final{Recent work~\cite{yu2024mavrl,kulkarni2024reinforcement} has shown the possibility of training sensorimotor controllers for obstacle avoidance end-to-end using reinforcement learning, paving the way towards a system that could solve drone racing completely end-to-end. However, these works still rely on explicit state estimation and a controller to execute velocity commands.}

Several other works apply a similar stacked perception and planning pipeline for other autonomous drone racing tasks~\cite{amer2021deep,loquercio2018dronet, gandhi2017learning, giusti2015machine}. We point the interested reader to existing surveys on the role of learning in drone navigation~\cite{Lee2021FlyingFA}.

A few works also studied the planning problem using data-driven methods, decoupling it from the perception problem.
An interesting approach demonstrated in the NeurIPS Game of Drones competition~\cite{Madaan2019AirsimDroneRacingLab} used an off-the-shelf reinforcement learning algorithm in place of a classic model-based planner for drone racing~\cite{ates2020long}.

\vspace*{-6pt}
\subsection{Learned Control}
\begin{figure}[h!]
    \centering
    \vspace*{-9pt}
    \begin{tikzpicture}[>=latex, font=\footnotesize]
    \tikzstyle{tb}=[minimum width = 1.4cm, text width = 1.2cm, minimum height = 0.4cm, align = center, draw, inner sep = 1pt, inner xsep=3pt];
    \newcommand\dx{0.4cm};
    \newcommand\dy{0.05cm};
    \newcommand\shift{0.05cm};
     \node [tb] (a) {Sensors};
     \node [tb, draw = none, above = \dy of a] (a2) {\emph{Hardware}};
     \node [tb, right = \dx of a] (b) {Perception};
     \node [tb, draw = none, above = \dy of b] (b2) {\emph{Software}};
     \node [tb, right = \dx of b] (c) {Planning};
     \node [tb, draw = none, right = \dx of c] (d) {};
     \node [tb, right = \dx of d] (e) {Drone};
     \node [tb, draw = none, above = \dy of d] (d2) {};
     \node [tb, draw = none, above = \dy of e] (e2) {};
     \draw [->] (a) -- (b);
     \draw [->] (b) -- (c);
     \draw [->] (c) -- (d);
     \draw [->] (d) -- (e);
     \draw (d2.north west) rectangle (d.south east) node [midway] {\includegraphics[width = 1.2cm]{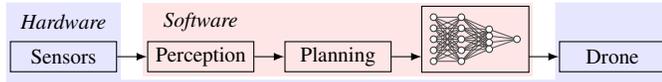}} ;
     \begin{pgfonlayer}{bg}    %
         \draw [draw=none, fill=red!10!white] ([xshift = -\shift, yshift=\shift]b2.north west) rectangle ([xshift=\shift, yshift=-\shift]d.south east);
         \draw [draw = none, fill=blue!10!white] ([xshift = -\shift, yshift=\shift]a2.north west) -| ([xshift=\shift, yshift=-2*\shift]a.south east) -| ([xshift=-\shift, yshift=\shift]e2.north west) -| ([xshift=\shift, yshift=-4*\shift]e.south east) -| ([xshift=\shift, yshift=-4*\shift]a.south east) -| cycle;
    \end{pgfonlayer}
    \end{tikzpicture}
    \vspace*{-8pt}
    \caption{Architecture 4: Learned Control}
    \label{fig:LearnedControl}
    \vspace*{-6pt}
\end{figure}

Data-driven control, like reinforcement learning, allows for overcoming many limitations of prior model-based controller designs by learning effective controllers directly from experience.
\change{For example, model-free RL was applied to low-level attitude control~\cite{koch2019reinforcement}, in which a learned low-level controller trained with PPO outperformed a fully tuned PID controller on almost every metric.
Similarly, \cite{lambert2019low}~used model-based RL for low-level control of an a priori unknown dynamic system.
More related to drone racing, recent works showcased the potential of learning-based controllers for high-speed trajectory tracking and drone racing\cite{kaufmann2022benchmark}.
Imitation learning is more data efficient compared to model-free RL.
In~\cite{li2020aggressive}, aggressive online control of a quadrotor has been achieved via training a network policy offline to imitate the control command produced by a model-based controller. Similarly, ~\cite{sanchez2018real} studied real-time optimal control via deep neural networks in an autonomous landing problem.
}
\final{Other work in this category has shown that reinforcement learning can find optimal~\cite{ferede2024end, song2021autonomous} or highly adaptive controllers~\cite{sacks2023deep}.}

With a learning-based controller, it can be difficult to provide robustness guarantees as with traditional methods such as the Linear Quadratic Regulator~(LQR).
While a learning-based controller may provide superior performance to classical methods \change{in simulation}, it may be that they cannot be used in the real world due to the inability to provide an analysis of the controller's stability properties. \change{This is particularly problematic for tracking the time-optimal trajectories required by drone racing.}
Recent works have attempted to address this using Lyapunov-stable neural network design for the control of quadrotors~\cite{dai2021lyapunov}.
This work shows that it is possible to have a learning-based controller with guarantees that can also outperform classical LQR methods.
Building upon this concept, reachability analysis, and safety checks can be embedded in a learned Safety Layer~\cite{selim2022safe}.

\vspace*{-6pt}
\subsection{Learned Planning \& Control}
\begin{figure}[h]
    \centering
    \vspace*{-9pt}
    \begin{tikzpicture}[>=latex, font=\footnotesize]
    \tikzstyle{tb}=[minimum width = 1.4cm, text width = 1.2cm, minimum height = 0.4cm, align = center, draw, inner sep = 1pt, inner xsep=3pt];
    \newcommand\dx{0.4cm};
    \newcommand\dy{0.05cm};
    \newcommand\shift{0.05cm};
     \node [tb] (a) {Sensors};
     \node [tb, draw = none, above = \dy of a] (a2) {\emph{Hardware}};
     \node [tb, right = \dx of a] (b) {Perception};
     \node [tb, draw = none, above = \dy of b] (b2) {\emph{Software}};
     \node [tb, draw = none, right = \dx of c] (d) {};
     \node [tb, right = \dx of d] (e) {Drone};
     \node [tb, draw = none, above = \dy of c] (c2) {};

     \node [tb, draw = none, above = \dy of d] (d2) {};
     \node [tb, draw = none, above = \dy of e] (e2) {};
     \draw [->] (a) -- (b);
     \draw [->] (b) -- (c);
     \draw [->] (d) -- (e);
     \draw (c2.north west) rectangle (d.south east) node [midway] {\includegraphics[width = 2.4cm]{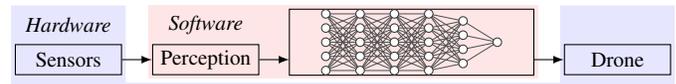}} ;
     \begin{pgfonlayer}{bg}    %
         \draw [draw=none, fill=red!10!white] ([xshift = -\shift, yshift=\shift]b2.north west) rectangle ([xshift=\shift, yshift=-\shift]d.south east);
         \draw [draw = none, fill=blue!10!white] ([xshift = -\shift, yshift=\shift]a2.north west) -| ([xshift=\shift, yshift=-2*\shift]a.south east) -| ([xshift=-\shift, yshift=\shift]e2.north west) -| ([xshift=\shift, yshift=-4*\shift]e.south east) -| ([xshift=\shift, yshift=-4*\shift]a.south east) -| cycle;
    \end{pgfonlayer}
    \end{tikzpicture}
    \vspace*{-8pt}
    \caption{Architecture 5: Learned Planning \& Control}
    \label{fig:LearnedPlanningAndControl}
    \vspace*{-6pt}
\end{figure}
 
The second paradigm of learned control is to produce the control command directly from state inputs without requiring a high-level trajectory planner, as shown in the architecture diagram of Figure~\ref{fig:LearnedPlanningAndControl}. 
\change{This approach enabled an autonomous drone with only onboard perception, for the first time, to outperform a professional human, and is state-of-the-art at the time of writing~\cite{kaufmann23champion}.}
In autonomous drone racing, this was proposed by~\cite{song2021autonomous, Song23Reaching}, where a neural network policy is trained with reinforcement learning to fly through a race track in simulation in near-minimum time.
Major advantages of the reinforcement-learning-based method are its capability to handle large track changes and the scalability to tackle large-scale random track layouts while retaining computational efficiency.
In~\cite{penicka2022learning}, deep reinforcement learning is combined with classical topological path planning to train robust neural network controllers for minimum-time quadrotor flight in cluttered environments. 
The learned policy solves the planning and control problem simultaneously, forgoing the need for explicit trajectory planning and control. 

\final{In this same category, another class of algorithms try to exploit the benefits of model-based and learning-based approaches using differentiable optimizers approaches~\cite{amos2018differentiable, theseus, pypose}, which leverage differentiability through controllers. 
For example, for tuning linear controllers by getting the analytic gradients \cite{difftune}, or  for creating a differentiable prediction, planning and controller pipeline for autonomous vehicles \cite{karkus2023diffstack}.
On this same direction, \cite{romero2024icra} equips the RL agent with a differentiable MPC \cite{amos2018differentiable}, located at the last layer of the actor network that provides the system with online replanning capabilities and allows the policy to predict and optimize the short-term consequences of its actions while retaining the benefits of RL training.
}

All these methods inherit the classic advantage of policy learning.
In addition, they do not require an external controller to track the plan. This eliminates the discrepancy between the planning and deployment stages, which is one of the main limitations of traditional planning methods (Sec.~\ref{sec:traditional_planning}).
Some of the limitations of traditional planning remain, such as the requirement of a globally-consistent state estimation and a map of the environment.
Also, they have not yet been demonstrated in sparse long-horizon planning problems, e.g. flying through a maze at high speeds, where their performance would likely drop due to sample complexity.

\vspace*{-6pt}
\subsection{End-to-End Flight}
\label{sec:end2end}
\begin{figure}[h]
    \centering
    \vspace*{-9pt}
    \begin{tikzpicture}[>=latex, font=\footnotesize]
    \tikzstyle{tb}=[minimum width = 1.4cm, text width = 1.2cm, minimum height = 0.4cm, align = center, draw, inner sep = 1pt, inner xsep=3pt];
    \newcommand\dx{0.4cm};
    \newcommand\dy{0.05cm};
    \newcommand\shift{0.05cm};
     \node [tb] (a) {Sensors};
     \node [tb, draw = none, above = \dy of a] (a2) {\emph{Hardware}};
     \node [tb, draw = none, right = \dx of a] (b) {};
     \node [tb, draw = none, above = \dy of b] (b2) {};
     \node [tb, draw = none, right = \dx of b] (c) {};
     \node [tb, draw = none, right = \dx of c] (d) {};
     \node [tb, right = \dx of d] (e) {Drone};
     \node [tb, draw = none, above = \dy of d] (d2) {\emph{Software}};
     \node [tb, draw = none, above = \dy of e] (e2) {};
     \draw [->] (a) -- (b);
     \draw [->] (c) -- (e);
     \draw (b2.north west) rectangle (c.south east) node [midway] {\includegraphics[width = 2.4cm]{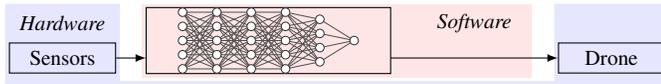}} ;
     \begin{pgfonlayer}{bg}    %
         \draw [draw=none, fill=red!10!white] ([xshift = -\shift, yshift=\shift]b2.north west) rectangle ([xshift=\shift, yshift=-\shift]d.south east);
         \draw [draw = none, fill=blue!10!white] ([xshift = -\shift, yshift=\shift]a2.north west) -| ([xshift=\shift, yshift=-2*\shift]a.south east) -| ([xshift=-\shift, yshift=\shift]e2.north west) -| ([xshift=\shift, yshift=-4*\shift]e.south east) -| ([xshift=\shift, yshift=-4*\shift]a.south east) -| cycle;
    \end{pgfonlayer}
    \end{tikzpicture}
    \vspace*{-8pt}
    \caption{Architecture 7: End to End Learning}
    \label{fig:end2end}
    \vspace*{-6pt}
\end{figure}

Expert pilots take raw sensory images from a first-person-view camera stream and map directly to control commands.
In this section, we explore approaches emulating this holistic navigation paradigm in autonomous drones.

Two families of approaches can be used to pursue an end-to-end navigation paradigm.
The first is substituting each of the perception, planning, and control blocks with a neural network.
This structure is followed by~\cite{li2018oil, muller2019learning}, where the authors train a perception-planning network and a control network using imitation learning.  
The perception network takes raw images as input and predicts waypoints to the next gate.
The control network uses such predictions with ground-truth velocity and attitude information to predict control commands for tracking the waypoints.
They showed improvements over pure end-to-end approaches, which directly map pixels to control commands and were able to show competitive lap times on par with intermediate human pilots within the Sim4CV simulator~\cite{muller2018sim4cv}.
Yet, the division into independent blocks leads to compounding errors and latencies, which negatively affect performance when flying at high speeds~\cite{Loquercio21FlightWild}. 

The second family of approaches directly maps sensor observation to commands without any modularity.
This design is used by~\cite{muller2018teaching}, which to date remains the only example of the completely end-to-end racing system.
Indeed, other end-to-end systems generally require an inner-loop controller and inertial information to be executed.
For instance, \cite{rojas2020deeppilot} trains an end-to-end CNN to directly predict roll, pitch, yaw, and altitude from camera images.
Similarly, \cite{fu2022learning, xing2023contrastive}~use a neural network to predict commands directly from vision.
To improve sample complexity, they use contrastive learning to extract robust feature representations from images and leverage a two-stage learning-by-cheating framework.

Independently of the design paradigm they follow, end-to-end navigation algorithms are currently bound to simulation.
The reasons why no method was successfully deployed in the real world include weak generalization to unseen environments, large computational complexity, and inferior performance to other modular methods.
Another interesting observation is that humans can pilot a drone exclusively
from visual observations.
Conversely, except for~\cite{muller2018teaching}, end-to-end systems still rely on the state extracted from other measurement modalities, e.g. an IMU.
The question of whether autonomous drones can race in the real world at high-speed without any inertial information remains open.
We provide more details on this question in Section~\ref{sec:challenges}.

\vspace*{-6pt}
\subsection{Discussion} \label{sec:LearningDiscussion}
Data-driven approaches are revolutionizing the research in autonomous drone racing, ranging from improving the system model to end-to-end control. 
Currently, the best-performing algorithms for drone racing include a learning-based component~\cite{foehn2022alphapilot,de2021artificial}, and this trend is unlikely to change in the coming years.
Indeed, compared to classical model-driven design, they can process high-dimensional sensory inputs directly, can be made robust to any modeling uncertainty (e.g. latency) by simply incorporating it in the training pipeline, and require far less engineering effort for tuning and deploying them~\cite{kaufmann2022benchmark}.

Our analysis shows that the majority of learning-based approaches heavily rely on simulators. While simulators may get better and faster in the near future, recent advances in real-world training~\cite{Wu22Day,smith22walk} and fine-tuning~\cite{smith2022legged,loquercio2022learn} offer a potential alternative for zero-shot simulation to reality transfer for sensorimotor policies.
However, so far, these works have been limited to legged locomotion.
Extension to agile drones could lead to the successful deployment of end-to-end policies, possibly improving the state of the art in agile flight.

\change{
Another limitation of the approaches discussed in this section is their inability to adapt to new and uncertain environments quickly.
The field of adaptive control has studied this problem extensively\cite{ioannou2012robust, aastrom2013adaptive, lavretsky2013robust}. Inspired by these works, there has been a recent push to use advancements in machine learning within the adaptive control framework.
A method to learn parametric uncertainty functions is introduced in~\cite{richards2021adaptive}.
These uncertainty functions could be learned offline using data captured from agile flight experiments, and then embedded within an adaptive controller to adjust controller parameters online during flight.
Results indicate that highly accurate trajectory tracking can be achieved with this approach, even in the face of strong wing gusts exceeding \unit[6.5]{m/s}.
More recently, learning-based controllers have shown the ability to adapt zero-shot to large variations in hardware and external disturbances~\cite{Dingqi22Adaptive}.
We see this as a promising area of research and one that is integral for reliable performance in changing environmental conditions.}

\section{Drone Racing Simulators}\label{sec:simulators}
One tool that has drastically accelerated the progress of research in autonomous drone flight is the use of simulation environments that attempt to recreate the conditions that real drones experience when flying.
Over the years, several simulation environments have been developed for the use of general research.

In 2016, the widely used RotorS simulation environment was published, which extends the capabilities of the popular Gazebo simulation engine to multi-rotors~\cite{furrer2016rotors}.
Gazebo uses the Bullet physics engine for basic dynamic simulation and contact forces.
Linear drag on the body of the multicopter is simulated based on the cross-sectional area and linear velocity of the simulated object. 
The RotorS extension features many easy-to-use plugins for developing multi-rotors, however, it distinctly lacks the photorealistic details needed to simulate accurate behavior of estimation and perception pipelines.

AirSim was introduced by Microsoft in 2018 as a photo-realistic simulator for the control of drones~\cite{shah2018airsim}.
It is built on the Unreal graphics engine and features easy-to-use plugins for popular flight controllers such as PX4\footnote{https://px4.io/}, ArduPilot\footnote{https://ardupilot.org/}, and others.
It was used in the 2019 NeurIPS Game of Drones challenge~\cite{Madaan2019AirsimDroneRacingLab}.
Because of the photorealism of AirSim, it is possible to simulate the entire perception and estimation pipeline with a good possibility of transfer to real-world drone systems.
Additionally, AirSim comes pre-packaged with an OpenAI-Gym environment for training Reinforcement Learning algorithms.
Organizations such as Bell, Airtonomy, and NASA are using AirSim to generate training data for learning-based perception models.

FlightGoggles~\cite{Guerra2019flightgoggles} was developed as another photorealistic simulator and was used as the primary simulation environment for the Lockheed Martin AlphaPilot challenge.
FlightGoggles contains two separate components: a photorealistic rendering engine
built with Unity3D and a dynamic simulation implemented in C++. 
FlightGoggles provides an interface with real-world vehicles using a motion capture system; such an interface allows the rendering of simulated images that correspond to the position of physical vehicles in the real world.

A recent simulator focused on Safe RL was proposed in~\cite{Panerati2021learning}.
It uses Gazebo and the Pybullet physics engine as the backend.
Leaderboards for several safety-focused training environments exist, encouraging researchers to submit their approaches and compete with other researchers around the world.

Flightmare~\cite{song2020flightmare} is a simulation environment featuring photorealistic graphics provided by the Unity engine.
The physics engine is decoupled and can be swapped out with various engines for user-defined levels of simulation fidelity.
Similar to FlightGoggles, Flightmare can also provide hardware-in-the-loop simulation functions where a virtual, synthetic camera image can be provided to the drone for use in control and estimation~\cite{Foehn22Agi}.

\final{Finally, Aerial Gym~\cite{kulkarni2023aerial} is a GPU-accelerated simulator that allows simulating millions of multirotor vehicles in parallel with nonlinear geometric controllers for attitude, velocity and position tracking. 
Additionally, the simulator offers a flexible interface for modeling a large number of obstacles and generating data such as RGB, depth, segmentation, and optical flow. }

\section{Competitions}\label{sec:competitions}
To gauge the progress of the field as a whole, several drone racing competitions have taken place since 2016.
We include a graphical overview of these events in Figure \ref{fig:timeline}.
The Autonomous Drone Racing (ADR) competition was an annual competition which took place during the IROS conference between 2016 and 2019.
In 2016, 11 teams competed in autonomous drone racing and were tasked to navigate a series of gates in sequence.
The positions of the gates were not known to the participating teams ahead of time, therefore teams flew very cautiously identifying the next waypoints online.
Each team was given 30 minutes prior to the official competition to fly the course as many times as they wished.
The winning team, from KAIST, made it through 10 of the 26 gates in 1 minute and 26 seconds.
For comparison, a human was able to complete the entire 26-gate course in 1 minute 31 seconds.
A survey summarizing the approaches used for these early competitions can be found in~\cite{moon2017iros}.
The following year, a similar competition took place during IROS in Vancouver, Canada, with better results.
This time, 14 teams participated and were given a CAD drawing of the course prior to the event with locations and dimensions of all gates.
Only 5 teams participated in the final in-person event, with the winning team making it through 9 out of 13 gates in over 3 minutes.
A summary of the winning approaches can be found in~\cite{moon2019challenges}.
Two more ADR competitions took place at IROS 2018 and 2019, with drones navigating courses faster and more reliably.

In 2019, Lockheed Martin sponsored the AlphaPilot AI Drone Racing Innovation Challenge where a 1 million dollar grand prize was awarded to the winning team~\cite{lockheedmartin2020alphapilot}.
The competition took place first in a virtual qualifying round which used the FlightGoggles simulation environment~\cite{Guerra2019flightgoggles}.
Nine teams out of more than 400 worldwide qualified for the final challenge which included navigating a new track in a time-trial setting against an expert human pilot.
\change{Such competition took the form of a tournament, with three
seasonal races and a final championship race. This made it very different from previous single-day competitions.
}
Ultimately, professional pilot Gabriel Kocher, from the Drone Racing League, manually piloted his drone through the course in only 6 seconds.
It took 11 seconds to the winner, MAVLab from TU Delft, and 15 seconds to the second-place winner, UZH-RPG from the University of Zurich, to complete the course autonomously.
The two different approaches are documented in \cite{de2021artificial, foehn2022alphapilot}.
Further comments are provided by the winner in  \cite{Wagter2021LearningFI}.
Perez et al. provide an overview of the types of hardware used for some of the drone racing competitions mentioned so far \cite{perez2021onboard}.

In 2019, the Game of Drones competition took place at the NeurIPS conference.
This competition was purely simulation-based and used the AirSim simulation environment built by Microsoft~\cite{microsoft2019gameofdrones,shah2018airsim,Madaan2019AirsimDroneRacingLab}.
Participants in the Game of Drones competition raced against simulated opponents in a head-to-head fashion, similar to how humans compete in FPV drone racing. 
Teams raced against a single simulated opponent, navigating through a complex series of gates in three different tiers: Planning Only, Perception Only, and Perception with Planning.

In 2022, at the Swiss Drone Days event in Zurich, Switzerland, three of the world's best human pilots competed against researchers from the Robotics and Perception Group of the University of Zurich. 
Flight speeds exceeding~\unit[100]{kph} were demonstrated by the autonomous drones.
When relying on motion capture, the autonomous drones were able to achieve significantly faster laptimes than the expert human pilots.
They additionally demonstrated it was possible to win races without motion capture, using only onboard computing and sensors to navigate the race track.
IEEE Spectrum author Evan Ackermann discusses the multi-day event in~\cite{ackerman_2022}.

\final{Looking into the future, the Abu Dhabi Autonomous Racing League  recently announced plans for an autonomous drone racing competition in 2025.}

\section{Datasets, Hardware, and Open Source Code}\label{sec:datasets}
In this section, we provide an overview of the existing open source code bases, useful datasets for autonomous drone racing \change{as well as hardware considerations}. 
We first discuss datasets, and then group the existing open source code bases by their use-cases in table \ref{table-example} and \change{conclude with a brief overview over drone racing hardware}.
\begin{table*}[!htb]
\centering
\footnotesize
\caption{Open Source Software and Datasets}
\setlength{\tabcolsep}{2pt}
\begin{tabularx}{1\linewidth}{C|c|c|c}
    \toprule
    \textbf{Name and Reference} &  \textbf{Category} & \textbf{Year} & \textbf{Link}\\
    \hline
    PAMPC \cite{falanga2018pampc} & Controller & 2018 & \url{https://github.com/uzh-rpg/rpg_mpc} \\
    \hline
    Deep Drone Acrobatics \cite{kaufmann2020deep} & Controller & 2019 & \url{https://github.com/uzh-rpg/deep_drone_acrobatics}\\
    \hline
    Data Driven MPC \cite{torrente2021datadriven} & Controller & 2020 & \url{https://github.com/uzh-rpg/data_driven_mpc} \\
    \hline
    High MPC \cite{song2022policy} & Controller & 2022 & \url{https://github.com/uzh-rpg/high_mpc} \\
    \hline
    AutoTune \cite{loquercio2022autotune} & Controller Tuner & 2022 & \url{https://github.com/uzh-rpg/mh_autotune} \\ 
    \hline
    Blackbird~\cite{antonini2018blackbird} & Dataset & 2018 & \url{https://github.com/mit-aera/Blackbird-Dataset} \\
    \hline
    UZH-FPV\cite{delmerico2019UZH} & Dataset & 2019 & \url{https://fpv.ifi.uzh.ch/} \\
    \hline
    NeuroBEM\cite{bauersfeld2021neurobem} &  Dataset & 2020 & \url{https://rpg.ifi.uzh.ch/NeuroBEM.html}\\
    \hline
    Eye Gaze Drone Racing \cite{pfeiffer2021human} & Dataset & 2021 & \url{https://osf.io/gvdse/}\\
    \hline
    TII Drone Racing Dataset~\cite{bosello2024race} & Dataset & 2024 & \url{https://github.com/tii-racing/drone-racing-dataset} \\
    \hline
    Time-optimal Planning for Quadrotor Waypoint Flight \cite{foehn2021time} & Planner & 2021 & \url{https://github.com/uzh-rpg/rpg_time_optimal} \\ 
    \hline
    Minimum-Time Quadrotor Waypoint Flight in Cluttered Environments\cite{penicka22RALmintimeplanning} & Planner & 2022 & \url{https://github.com/uzh-rpg/sb_min_time_quadrotor_planning} \\ 
    \hline
    RotorS \cite{furrer2016rotors} & Simulator & 2016 & \url{https://github.com/ethz-asl/rotors_simulator} \\
    \hline
    AirSim \cite{Madaan2019AirsimDroneRacingLab} & Simulator & 2018 & \url{https://microsoft.github.io/AirSim/} \\
    \hline
    FlightGoggles\cite{Guerra2019flightgoggles} & Simulator & 2019 & \url{https://github.com/mit-aera/FlightGoggles}\\ 
    \hline
    Flightmare\cite{song2020flightmare} & Simulator & 2020 & \url{https://uzh-rpg.github.io/flightmare/}\\ 
    \hline
    Learning to fly—a gym environment with pybullet physics for reinforcement learning of multi-agent quadcopter control\cite{Panerati2021learning} & Simulator & 2021 & \url{https://github.com/utiasDSL/gym-pybullet-drones}\\
    \hline
    Aerial Gym\cite{kulkarni2023aerial} & Simulator & 2023 & \url{https://github.com/ntnu-arl/aerial_gym_simulator} \\ 
    \hline
    Sim 2 Real Domain Randomization \cite{loquercio2019deep} & Sim2Real Transfer & 2019 & \url{https://github.com/uzh-rpg/sim2real_drone_racing} \\
    \hline
    RPG Quadrotor Control \cite{faessler2017differential} & Software Stack & 2017 & \url{https://github.com/uzh-rpg/rpg_quadrotor_control}\\
    \hline
    Agilicious \cite{Foehn22Agi} & Software Stack & 2022 & \url{https://github.com/uzh-rpg/agilicious}\\
    \hline
    Kalibr \cite{rehder2016extending} & Camera Calibration & 2022 & \url{https://github.com/ethz-asl/kalibr} \\ 
    \hline
    \specialchange{VID-Fusion \cite{ding2021vid}} & Estimation & 2021 & \url{https://github.com/ZJU-FAST-Lab/VID-Fusion}\\
    \hline
    \specialchange{Fast-Racing \cite{Han2021Fast-racing}} & Planner & 2021 & \url{https://github.com/ZJU-FAST-Lab/Fast-Racing}\\
    \hline
    \specialchange{Ego-planner \cite{egoplanner}} & Planner & 2021 & \url{https://github.com/ZJU-FAST-Lab/ego-planner}\\
    \hline
    \specialchange{GCOPTER \cite{WANG2022GCOPTER}} & Planner & 2022 & \url{https://github.com/ZJU-FAST-Lab/GCOPTER}\\
    \hline
    \specialchange{FASTER \cite{tordesillas2021faster}} & Planner & 2021 & \url{https://github.com/mit-acl/faster}\\
    \hline
    \specialchange{Panther \cite{tordesillas2022panther}} & Planner & 2022 & \url{https://github.com/mit-acl/panther}\\
    \hline
    \specialchange{Deep Panther \cite{tordesillas2023deeppanther}} & Planner & 2023 & \url{https://github.com/mit-acl/deep_panther}\\
    \hline
    \specialchange{Raptor \cite{zhou2021raptor}} & Planner & 2021 & \url{https://github.com/HKUST-Aerial-Robotics/Fast-Planner}\\
    \bottomrule
    
\end{tabularx}
\label{table-example}
\end{table*}

\subsection{Datasets}
In 2018, researchers from MIT released a large scale dataset for perception during aggressive UAV flight \cite{antonini2018blackbird}.
This dataset contains over 10 hours of flight data which includes simulated stereo and downward-facing camera images at~\unit[120]{Hz}, real-world IMU data at \unit[100]{Hz}, motor speed data at \unit[190]{Hz}, and motion capture data at \unit[360]{Hz}.
The sensor suite was chosen such that algorithms like Visual-Inertial Odometry (VIO) or Simultaneous Localization and Mapping (SLAM) could be evaluated on the dataset.

In 2019, the UZH-FPV Drone Racing Dataset was released, which contains many agile maneuvers flown by a professional racing pilot~\cite{delmerico2019UZH}.
The dataset includes indoors and outdoors real-world camera images, inertial measurements, event camera data, and ground truth poses provided by an advanced motion capture system (a total station) providing millimeter-level accuracy. \final{In 2024, the dataset was extended with new data recorded onboard an autonomous racing drone flying in a racing track with peak speed exceeding~\unit[20]{m/s}. This new data includes large field-of-view camera images, inertial measurements, and ground truth from a motion capture system.}
Similar to the authors in~\cite{antonini2018blackbird}, the authors of this dataset hope to push the state of the art in state estimation during aggressive motion and have created competitions to allow researchers to compete against one another on this agile flight benchmark.\footnote{\url{https://fpv.ifi.uzh.ch/uzh/uzh-fpv-leader-board/}}. \final{A recent effort reported in~\cite{bosello2024race} open-sourced high-quality data from both autonomous and human-piloted flights. This effort enables the study of both the perception and control problem without actual hardware, lowering the barrier of entry for studying drone racing.}

Research on how expert human pilots focus on their targets during flying and provide a dataset that contains flight trajectories, videos, and data from the pilots is examined in~\cite{pfeiffer2021human}

NeuroBEM~\cite{bauersfeld2021neurobem} is a hybrid aerodynamic quadrotor model which combines blade-element-momentum-theory models with learned aerodynamic representations from highly aggressive maneuvers.
While the model is fit to the specific quadrotor platform defined in~\cite{Foehn22Agi}, the approach can be used for any quadrotor platform and provides over 50\% reduction in model prediction errors compared to traditional, exclusively-first-principles approaches. 

\subsection{Open-Source Code}
A significant amount of autonomous drone racing research has been open sourced to the community, making implementation less daunting for newcomers to the field. 
A collection of all known drone racing repositories has been provided to the reader in Table \ref{table-example}.
These code bases range across controllers, planners, sensor calibration, and even entire software stacks dedicated to drone racing. 
We encourage both newcomers and experienced researchers to check out the extensive amount of open source code bases available and contribute back to the community.

\change{
\subsection{Hardware}
This survey does not intend to cover the hardware design of racing drones rigorously. For an in-depth overview, see~\cite{Foehn22Agi}, where the hardware and software design for developing a very capable research platform are discussed. To make this survey self-contained, this section presents a brief overview of the hardware design of a racing drone nevertheless.

\subsubsection{Racing Drone Design}
A suitable hardware design should maximize the agility and acceleration of the drone, and hence, it needs to be as lightweight as possible~\cite{kumar2012opportunities}. For drones featuring onboard compute, the drone size is thus lower-bounded by the size of the computer. Currently, the NVIDIA Jetson family is the smallest off-the-shelf hardware with sufficient compute to run complex neural networks, and it leads to drones built on \unit[6]{inch} frames. Carbon fiber offers an excellent compromise between the weight and durability of the frame, while other parts (such as holders for the computer) can be designed using a 3D printer.

For actuation, fast-spinning brushless DC motors are ideal because of their high specific power output, often exceeding \unit[500]{W} for a \unit[50]{g} motor. 
In general, larger propellers will improve the energy efficiency of the drone~\cite{bauersfeld22Range} while smaller propellers lead to a faster motor response. On a \unit[6]{inch} frame, three-bladed \unit[5]{inch} propellers present a good compromise. To sustain the power demand of brushless drone-racing motors (often exceeding \unit[2]{kW} at full throttle~\cite{bauersfeld22Range}) a lithium-polymer battery with a sufficiently high discharge current rating (e.g., \unit[120]{C}) is required.

The Pixhawk PX4 flightstack, despite being commonly used for quadrotors~\cite{mohta2018fast, Baca2021jirs}, fixed-wings~\cite{Meier2012pixhawk}, and hybrid VTOL platforms~\cite{Bauersfeld20VTOL}, is not optimized for agile flight. Conversely, agile autonomous research platforms~\cite{Foehn22Agi, kaufmann23champion} use Betaflight as a low-level controller, similar to professional human racing pilots.

The design of a capable racing drone is important for researchers developing new technology. However, in many drone racing competitions, the hardware design is not left to the participants but is standardized. This approach is common in human drone racing, where thousands of identical drones are built before each competition.
This concept was also adopted by the AlphaPilot~\cite{lockheedmartin2020alphapilot} competition, where all participants used a given platform. Overall, this approach ensures fair competition.

\subsubsection{Beyond Quadcopters}
While this survey focuses on multi-copter drones, future drone racing competitions will go beyond this platform. Indeed, FPV Fixed-Wing Racing is already a popular sport among human pilots~\cite{wingracing}.
For example, vertical takeoff and landing (VTOL) drones might offer a great alternative to quadcopters. VTOL aircraft combine the high speeds achieved by fixed-wing drones with some of the maneuverability of multicopters.
Pioneering works on this platform have already shown agile control~\cite{tal2022vtol} and trajectory generation for aerobatic VTOL flight~\cite{tal2023vtol}.
Perhaps, once such research platforms are available off the shelf, VTOL aircraft racing will become a popular platform for autonomous drone racing research.
}

\section{Open Research Questions and Challenges}\label{sec:challenges}

\change{While a lot of progress has been made, there are still many challenges to be overcome in drone racing research. In the following, we discuss the most interesting challenges in detail.}

\subsection{Challenge 1: Reliable State Estimation at High-Speeds}
In its current form, online, robust, and accurate state estimation is highly beneficial when pushing autonomous drones to their limits.
Currently, classical state estimation approaches based on visual-inertial odometry cannot cope with the perceptual challenges present in drone racing tasks.
Motion blur, low texture, and high dynamic range are some reasons why classical VIO algorithms accumulate large errors in localization.
The miscalibration of intrinsic and extrinsic camera parameters can lead to improper estimates of the camera pose on a drone. 
This is due to local movements of the camera frame relative to the drone body, as well as changes in temperature and pressure. 
VIO drift can render the state estimates unusable unless corrected through localizations to a prior map. 
New sensor modalities, such as event cameras, could potentially alleviate this issue. 
Although event-aided VIO algorithms for drones have been proposed to improve robustness to motion blur, they have not been demonstrated at high speeds as seen in drone racing. 
Future research in agile flight may focus on finding new event representations that are computationally efficient and compatible with classical VIO formulations.
One example is to exploit direct methods~\cite{hidalgo2022event}.
Other promising sensor modalities are motor speed controllers and force sensors.
These sensor measurements could be used to include more advanced drone models in VIO, e.g. modeling aerodynamics effects, in order to limit the drift that accumulates where camera measurements are degraded.
One of the main consequences of motion blur, low texture, and high dynamic range is unreliable feature extraction and matching.
This consequently degrades the performance of the visual frontend.
Deep learning methods have the potential to solve this problem.
What hinders the application of these methods to drone racing at the moment is their computational cost.
Future research should work on lightweight neural networks that can provide inference at a high rate.
Neural networks could also be used to remove non-zero mean noise and constant errors from the inertial measurements.
A potentially fruitful area of research is in combining neural networks for input processing with a geometry-based VIO backend.
This could lead to the next step in the research on VIO for drone racing.
Current works~\cite{teed2021droid, teed2022deep} have shown that this direction outperforms end-to-end visual-based odometry methods.

\vspace*{-8pt}
\subsection{Challenge 2: Flying from Purely Vision }

State-of-the-art autonomous navigation methods rely on visual and inertial information, usually combined with classic perception algorithms.
Conversely, expert human pilots rely on nothing more than a first-person-view video stream, which they use to identify goals and estimate the ego-motion of the drone.
Building systems that, similarly to human pilots, only rely on visual information is very interesting from a scientific perspective.
Indeed, since simulating RGB is very challenging, solving this question might require lifelong learning algorithms operating in the real world.
In addition, eliminating inertial information might have some engineering advantages too, e.g., data throughput, power consumption, and lower cost.
Seminal works in this direction try to understand how humans solve this task~\cite{pfeiffer2021human, pfeiffer2022visual}.
They found that expert pilots can control drones despite a 200ms latency, which is compensated by the human brain.
Taking inspiration from biology, a recent work~\cite{de2022accommodating} shows that it is possible to fly with camera images and an onboard gyroscope (e.g., removing the accelerometer), as long as the system never hovers.
However, the above questions still remain mostly open and a good avenue for research at the intersection of computer vision, neuroscience, and biology.

\vspace*{-8pt}
\subsection{Challenge 3: Multiplayer Racing}
Much of the work done up until this point on autonomous drone racing has focused on time-optimal flight without considering how a capable opponent might impact the competition dynamics.
In FPV races, pilots can compete against up to 5 opponents simultaneously, bringing about the need to anticipate how their opponents might behave. 
Humans are astonishingly capable of recognizing opportunities for overtaking and executing complex maneuvers in the face of large aerodynamic disturbances caused by flying close to another drone.
Achieving such capabilities requires an agent to estimate their opponent's state using only onboard visual sensors.
However, these observations in drone racing are sparse because the camera faces forward along the heading axis, meaning that the only time an opponent is observable is when the ego-agent is behind them.
Sophisticated motion and planning models which can propagate predictions of the opponents' states and racing lines through time are necessary to anticipate collisions or overtaking opportunities. 
One way to simplify the problem is combining classical vision with learning-based control, which has shown promising results in multi-agent zero-sum games for locomotion~\cite{bajcsy2023learning}.
An initial study~\cite{spica2020real} examined how game-theoretic planners can lead to highly competitive behavior in two-player drone racing, however, this work was confined to racing on a 2D plane.
The work was further extended to 3D spaces in~\cite{wang2020multi}, but there is a significant opportunity for researchers to explore the competitive nature of drone racing and develop interesting racing strategies that lead to time-optimal agents that are able to deal with complex opponent behavior.

\vspace*{-8pt}
\subsection{Challenge 4: Safety}
\change{Autonomous drone racing research has so far focused on demonstrating that superhuman performance in racing is possible in controlled conditions~\cite{kaufmann23champion} but has put less emphasis on risk and safety. 
We predict that this trend will soon change.
Adding safety to agile flight has gained much attention recently~\cite{Panerati2021learning, brunke2022safe}.
Initial works focused on generating a collision-free trajectory~\cite{chen2015realtime, gao2017gradient, gao2018online} with less emphasis on performance. More geared towards agile flight, the works~\cite{chow2015trading, singh2017robust, singh2020robust, tordesillas2021faster} have studied the problem of trading off safety and performance. All the aforementioned works rely on solving constrained optimization problems.
Outside of drone racing, similar paradigms have been developed and have the potential to inspire future algorithms. Such methods are, for example, conformal analysis~\cite{luo2022sample}, chance-constrained dynamic programming~\cite{ono2015chance}, control barrier functions~\cite{ames2019control}, or reachability analysis~\cite{bansal2017hamilton}. The latter has been successfully applied in the context of autonomous driving with collision avoidance~\cite{wang2020infusing, leung2020infusing}. 

More modern, learning-based methods have been explored for risk-aware autonomous driving in the context a of map-prediction approach~\cite{elhafsi2020map} and in combination with Tube MPC~\cite{wabersich2021probabilistic}, a form of MPC that takes stochasticity into account.
However, such approaches generally do not scale to high-dimensional perception but rely on robust state-estimation for all involved agents.
Combining such algorithms with the methods for vision-based, high-speed drone racing presented in this survey could solve both of these problems simultaneously. 
As a first step in this direction, recent work~\cite{bauersfeld2023conditioning} has shown that a learned control policy can be conditioned on an auxiliary input signal from a user. The signal regulates the maximally available thrust, leading to a single learned policy that can race at various speeds and risk levels.}

\vspace*{-8pt}
\subsection{Challenge 5: Transfer to Real-World Applications}
Drone racing, while an extraordinarily challenging research environment, is ultimately not the end goal.
Opportunities exist for technology transfer between the drone racing research community to real-world applications such as search and rescue, inspection, agriculture, videography, delivery, passenger air vehicles, law enforcement, and defense.
However, applications that leverage the full agility of the platform have much to gain. Drones that fly fast, fly farther, therefore increasing the productivity of drones in every commercial sector~\cite{bauersfeld22Range}.

\change{One of the major challenges to real-world application is generalization to conditions where the environmental knowledge before deployment is limited.
For example, we often do not have a known map ahead of time for real-world applications, which requires simultaneous estimation of the state of the drone while mapping the environment.
However, a central theme of drone racing research has been the development of adaptive control strategies and decision-making algorithms to enable drones to react rapidly to changes in the race track or the robot condition(Sec.~\ref{sec:control} and Sec.~\ref{sec:classic_perception}). These strategies can be used to handle real-world applications where environmental knowledge is imperfect and to enable adaptation to unforeseen obstacles and challenges. In addition, learning-based sensorimotor controllers for drones, increasingly more popular due to research on racing, have been designed with the ability to generalize from limited data, adapt, and improve their performance over time (Sec.~\ref{Learning-based-Approaches}). Such generalization and adaptation abilities have already been applied to cases where there is no previous knowledge of the environment~\cite{Loquercio21FlightWild}.}

Building algorithms that can continually improve from their experience is another alternative to favor this transfer.
While recent advances in reinforcement learning research point to the feasibility of this path~\cite{yu2019meta,loquercio2022learn,smith2022legged}, it is unclear when and how such recent approaches would be applicable to drones or similarly agile platforms in the real world.
Collecting data for continual RL onboard a drone is notoriously difficult.
This is because the drone does not have the luxury of remaining in contact with the ground like legged robots and cars, and thus has to immediately know how to hover otherwise a crash will occur.
One interesting area that may be useful for continual RL in drones is the notion of ``safe-RL".
The goal of safe RL is to enable exploration without ever incurring catastrophic failure of the system. 
Initial work on this topic can be found in~\cite{turchetta2020safe}.
A survey paper covering safe RL methods can be found in~\cite{brunke2022safe}.
Furthermore, a thorough review paper on continual, or life-long RL can be found in~\cite{khetarpal2020towards}.

\section{Conclusions and Summary}\label{sec:conclusions}
\change{
 From racing at a pace comparable to walking speed~\cite{moon2017iros}, autonomous drones have advanced to surpassing world champions~\cite {kaufmann23champion}.
 Such an exponential advance has been driven by both algorithmic innovations, e.g., learning sensorimotor controllers in simulation, and system engineering improvements. Such advances span the entire navigation pipeline: perception, planning, and control. Our paper comprehensively covers each of these topics. Methodologically, the dominant trend has been a shift from conventional methods to data-driven solutions. However, in contrast to fields like computer vision and natural language processing, neural networks did not replace but coexist with traditional methods: no method with competitive performance in the real world is fully data-driven. The most resilient part of the pipeline is state estimation, where strong prior knowledge about the dynamics and environment are still needed to cope with the lack of sensorimotor data. In the short term, we predict that such a hybrid approach could be applied to other physical systems, e.g., autonomous ground vehicles and personal robots.
However, in the long term, we predict that, similarly to research in computer vision and natural language processing, neural networks will replace each part of the pipeline. This will require many innovations, e.g., computationally efficient architectures, offline pre-training strategies, and fast adaptation schemes to previously unseen conditions.
While autonomous drones are already superhuman in controlled scenarios, many challenges are yet to be solved to outperform human champions in official drone racing leagues and transfer the findings to real-world applications.
}

\section{Acknowledgements}
\change{The authors thank Manasi Muglikar for her valuable inputs on event-camera methods for state estimation and perception.}

\section{Contributions}\label{sec:contributions}
Drew Hanover initiated the idea of this paper, created the paper structure, and contributed to all sections of this paper while coordinating efforts amongst the co-authors.
Antonio Loquercio contributed to the paper structure and the learning-based sections.
Leonard Bauersfeld authored the Drone Modeling section and created the graphics seen throughout.
Angel Romero contributed to the Classical Planning and Control sections.
Giovanni Cioffi contributed to the Classical Perception and Challenges sections.
Yunlong Song contributed to the Simulators and Learning-Based Planning/Control sections\change{.}
Robert Penicka contributed to both Classical and Learning-Based Planning sections.
Elia Kaufmann contributed to the paper structure and throughout the Learning-Based sections.
Davide Scaramuzza contributed to the general paper structure and revised the paper thoroughly and critically.

\def\UrlBreaks{\do\/\do-}
{\footnotesize
\bibliographystyle{IEEEtran}
\bibliography{references}
}

\clearpage

\begin{IEEEbiography}[{\includegraphics[width=1in,height=1.25in,clip,keepaspectratio]{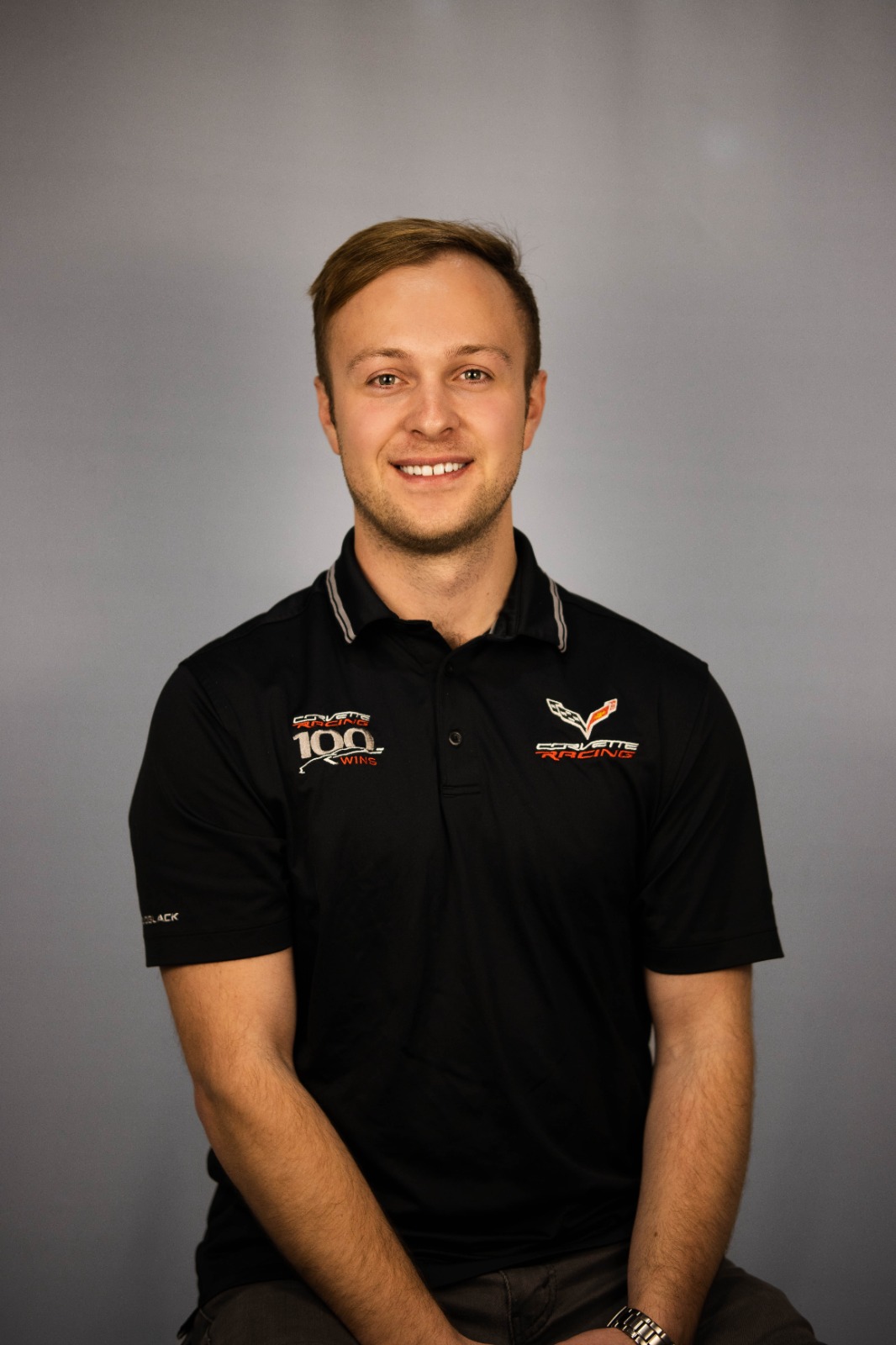}}]{Drew Hanover}  is the Chief Technology Officer and Founder of Innovire AG. He completed his Bachelors in Mechanical Engineering at Michigan Technological University, and his Masters in Robotics at the University of Michigan. He has spent time working with NASA, General Motors, and Pratt and Miller Engineering across a multitude of engineering domains.
\end{IEEEbiography}
\begin{IEEEbiography}[{\includegraphics[height=1.2in, keepaspectratio]{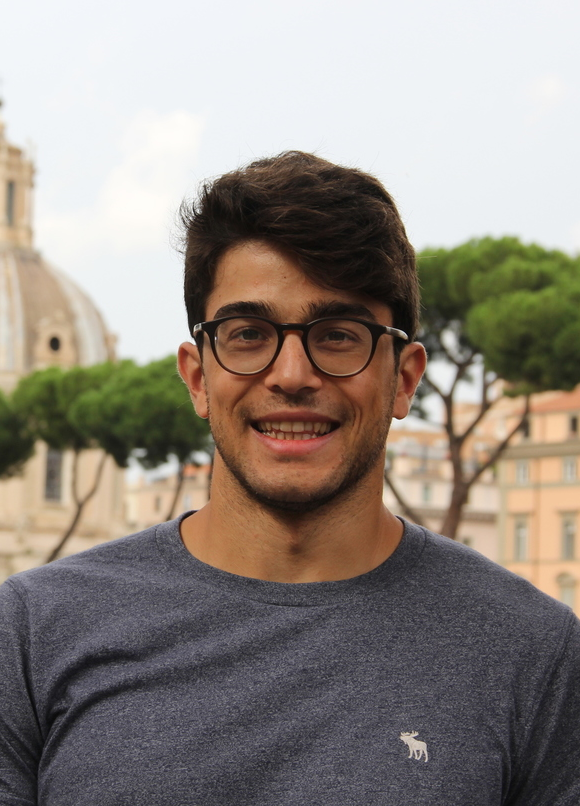}}]{Antonio Loquercio} is a  professor of electrical engineering and computer science at the University of Pennsylvania. He received a M.Sc. degree from ETH Zurich and a Ph.D. from the University of Zurich in 2017 and 2021, respectively. He worked at the Berkeley Artificial Intelligence Research Lab (BAIR) at UC Berkeley from 2022 to 2024.
\end{IEEEbiography}
\begin{IEEEbiography}[{\includegraphics[width=1in,height=1.25in,clip,keepaspectratio]{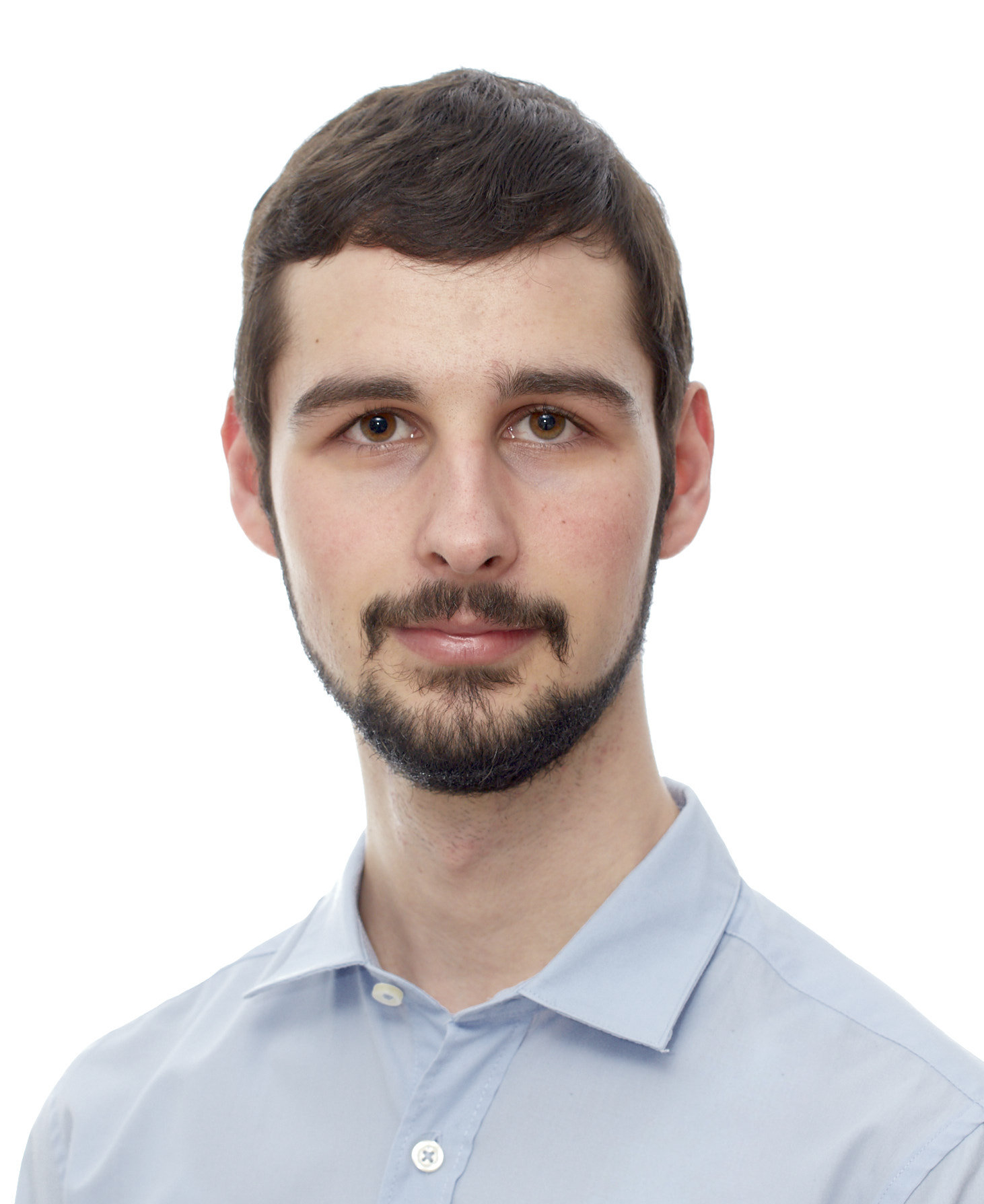}}]{Leonard Bauersfeld} received his M.Sc. degree in robotics, system and control from ETH Zurich, Switzerland
in 2020. He is currently a PhD student in the Robotics and Perception Group at the University of Zurich, led by Prof. Davide Scaramuzza. His reseach interests are autonomous vision-based quadrotor flight and quadrotor simulations. He works novel approaches, combining first-principles methods with modern data-driven models to advance agile quadrotor flight. 
\end{IEEEbiography}
\begin{IEEEbiography}[{\includegraphics[width=1in,height=1.25in,clip,keepaspectratio]{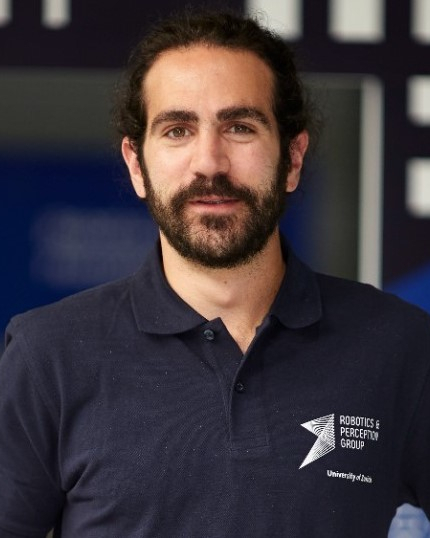}}]{Angel Romero} received a MSc degree in "Robotics, Systems and Control" from ETH Zurich in 2018.
Previously, he received a B.Sc. degree in Electronics Engineering from the University of Malaga in 2015.
He is currently working toward a Ph.D. degree in the Robotics and Perception Group at the University of Zurich, finding new limits in the intersection of machine learning, optimal control, and computer vision applied to super agile autonomous quadrotor flight under the supervision of Prof. Davide Scaramuzza.
\end{IEEEbiography}
\begin{IEEEbiography}[{\includegraphics[width=1in,height=1.25in,clip,keepaspectratio]{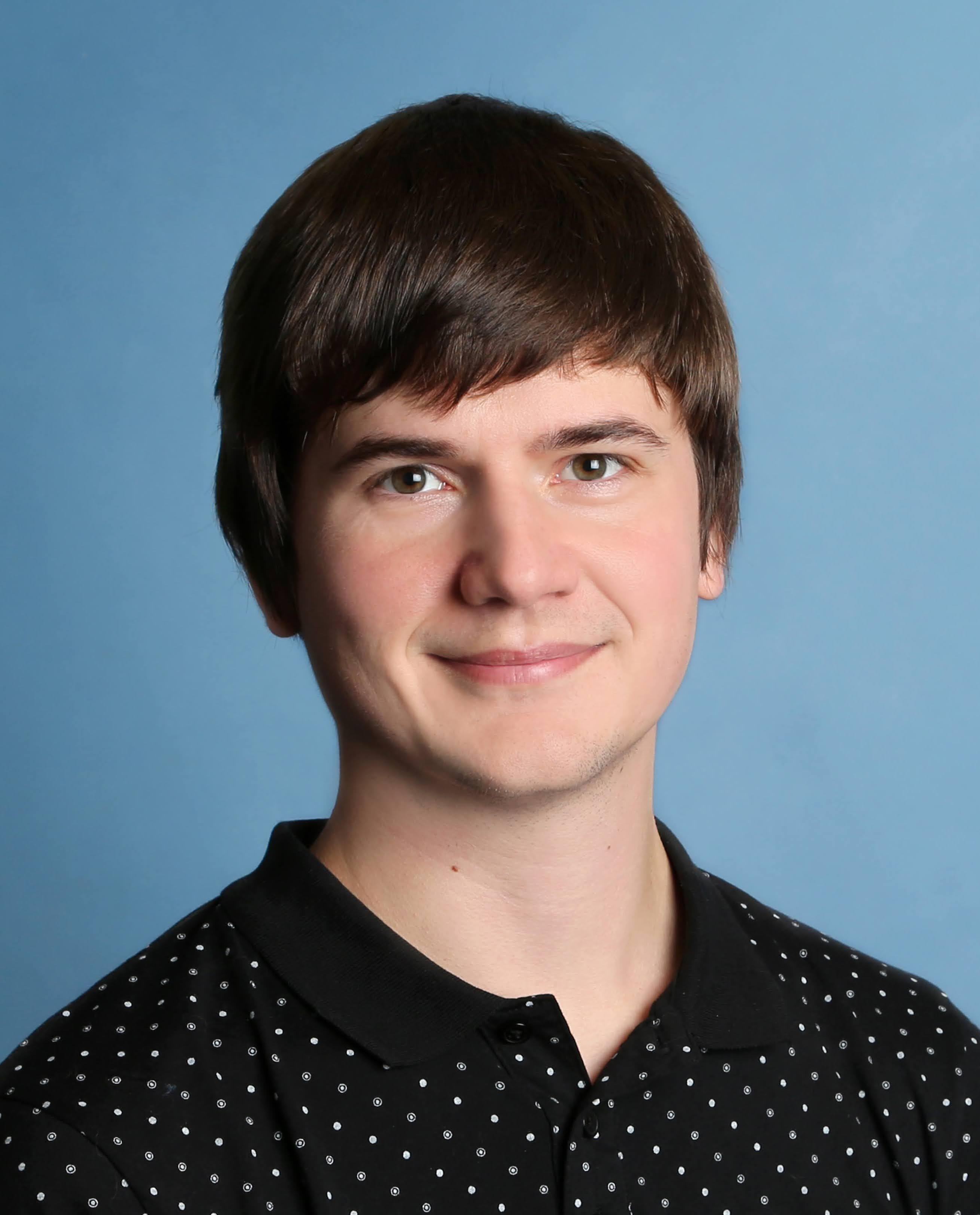}}]{Robert Penicka}  
is currently a postdoc in the Multi-Robot Systems (MRS) group at the  Czech Technical University (CTU) in Prague. He did his Ph.D. at the CTU in Prague in 2020 and was a postdoctoral researcher at the University of Zurich between 2020 and 2022 under the supervision of Professor Scaramuzza. Since 2022, he's been a research fellow at CTU, focusing on high-level mission planning, trajectory planning, and control for UAVs. He's bridged the gap between mission planning and trajectory planning, particularly in cluttered environments, earning recognition including the Dean’s Prize and 2nd place in the Werner von Siemens Award for Industry 4.0. He's also won the Joseph Fourier Prize and the Antonin Svoboda Award for his doctoral thesis.
\end{IEEEbiography}
\begin{IEEEbiography}[{\includegraphics[height=1.2in, keepaspectratio]{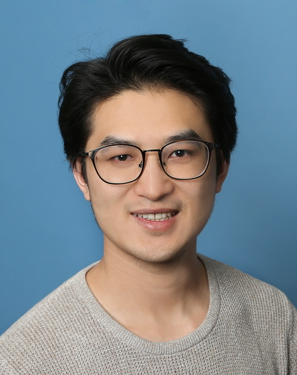}}]{Yunlong Song} obtained the M.Sc. degree in Information and Communication Engineering from Technical University of Darmstadt in 2018. He is currently a Ph.D. student in the Robotics and Perception Group at the University of Zurich under the supervision of Prof. Davide Scaramuzza.
His research interests include reinforcement learning, machine learning, and robotics.
\end{IEEEbiography} 
\begin{IEEEbiography}[{\includegraphics[width=1in,height=1.25in,clip,keepaspectratio]{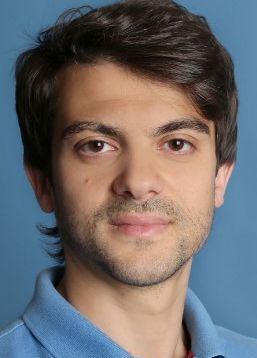}}]{Giovanni Cioffi} %
holds an M.Sc. in Mechanical Engineering from ETH Zürich, Switzerland, which he obtained in 2019. He is currently pursuing a Ph.D. at the University of Zürich under the supervision of Prof. Davide Scaramuzza. His research centers on the intersection of computer vision and robotics, exploring topics such as visual(-inertial) odometry and SLAM. His contributions were recognized by multiple awards in top-tier robotic conferences and journals, such as the IROS 2023 Best Paper Award and the RA-L 2021 Best Paper Award.
\end{IEEEbiography}
\begin{IEEEbiography}[{\includegraphics[width=1in,height=1.25in,clip,keepaspectratio]{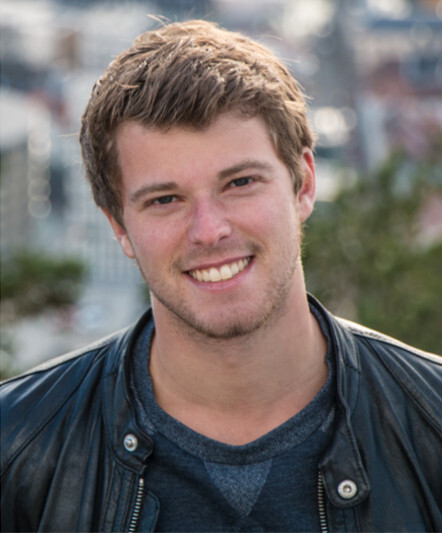}}]
{Elia Kaufmann} completed his Ph.D. in Informatics at the University of Zurich in 2022, where he was supervised by Prof. Davide Scaramuzza. His doctoral research focused on advancing the application of machine learning techniques to enhance perception and control of autonomous aerial vehicles. He earned an M.Sc. degree in Robotics, Systems, and Control from ETH Zurich in 2017, after obtaining a B.Sc. in Mechanical Engineering in 2014. Currently, he is a Senior Autonomy Engineer at Skydio.
\end{IEEEbiography}
\begin{IEEEbiography}[{\includegraphics[width=1in,height=1.25in,clip,keepaspectratio]{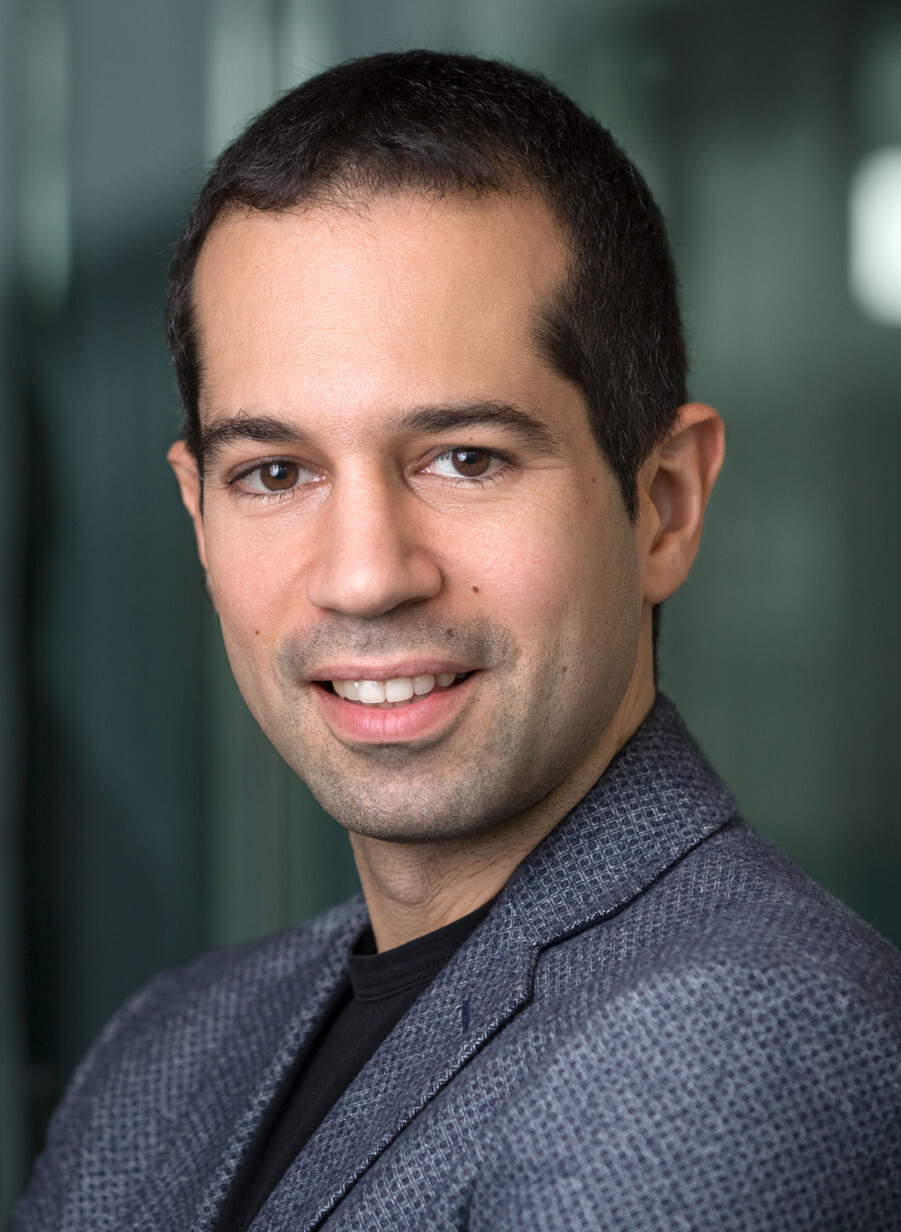}}]{Davide Scaramuzza} %
is a Professor of Robotics and Perception at the University of Zurich. He did his Ph.D. at ETH Zurich, a postdoc at the University of Pennsylvania, and was a visiting professor at Stanford University. His research focuses on autonomous, agile microdrone navigation using standard and event-based cameras. He pioneered autonomous, vision-based navigation of drones, which inspired the navigation algorithm of the NASA Mars helicopter and many drone companies. He contributed significantly to visual-inertial state estimation, vision-based agile navigation of microdrones, and low-latency, robust perception with event cameras, which were transferred to many products, from drones to automobiles, cameras, AR/VR headsets, and mobile devices. In 2022, his team demonstrated that an AI-controlled, vision-based drone could outperform the world champions of drone racing, a result that was published in Nature. He is a consultant for the United Nations on disaster response, AI for good, and disarmament. He has won many awards, including an IEEE Technical Field Award, the IEEE Robotics and Automation Society Early Career Award, a European Research Council Consolidator Grant, a Google Research Award, two NASA TechBrief Awards, and many paper awards. In 2015, he co-founded Zurich-Eye, today Meta Zurich, which developed the world-leading virtual-reality headset Meta Quest. In 2020, he co-founded SUIND, which builds autonomous drones for precision agriculture. Many aspects of his research have been featured in the media, such as The New York Times, The Economist, and Forbes.
\end{IEEEbiography}

\end{document}